\documentclass{article}

\usepackage{amsfonts}
\usepackage{amsmath}
\usepackage{amssymb}
\usepackage{amsthm}
\usepackage{geometry}
\usepackage{mathtools}
\usepackage{graphicx}
\usepackage{subfigure}
\usepackage{lineno}
\usepackage{hyperref}
\modulolinenumbers[1]
\usepackage{mathrsfs}
\usepackage{url}
\usepackage{xcolor}
\usepackage{xspace}
\usepackage{stmaryrd}
\usepackage[ruled, linesnumbered, vlined]{algorithm2e}
\usepackage{multirow}
\usepackage{rotating}
\usepackage{multicol}
\usepackage{ authblk }

\newtheorem{theorem}{Theorem}
\newtheorem{example}[theorem]{Example}

\newif\iftodo
\todotrue  

\newcommand{\onto}[1]{\ensuremath{\cp{#1}}}

\newcommand{\rass}[3]{\mbox{$#3 (#1,#2)$}}

\newcommand{\cp}[1]{\mathsf{#1}}
\newcommand{\role}[1]{\mathsf{#1}}

\newcommand{\ind}[1]{\mathsf{#1}}

\newcommand{\alc}{\ensuremath{\mathcal{ALC}}}

\newcommand{\all}{\forall}
\newcommand{\andc}{\sqcap}

\newcommand{\compr}{\circ}

\newcommand{\bottomc}{\perp}

\newcommand{\CASE}[1]{\STATE \textbf{case} #1\textbf{:} \begin{ALC@g}}

\newcommand{\cass}[2]{\mbox{$#2(#1)$}}

\newcommand{\csome}{\exists}

\newcommand{\DEFAULT}{\STATE \textbf{default:} \begin{ALC@g}}
\newcommand{\DEFAULTLINE}[1]{\STATE \textbf{default:} }

\newcommand{\eg}{e.g.}

\newcommand{\ENDCASE}{\end{ALC@g}}
\newcommand{\ENDDEFAULT}{\end{ALC@g}}

\newcommand{\highi}[1]{{#1}^\mathcal{I} }
\newcommand{\I}{\mathcal{I}}
\newcommand{\ie}{\textit{i.e.}}

\newcommand{\impc}{\sqsubseteq}

\newcommand{\K}{\ensuremath{\mathcal{K}}}

\newcommand{\KB}{\K}

\newcommand{\nd}{\noindent}

\newcommand{\notc}{\neg}

\renewcommand{\O}{\ensuremath{\mathcal{O}}}
\newcommand{\oneof}[1]{\{#1 \}}

\newcommand{\orc}{\sqcup}

\newcommand{\some}{\exists}

\newcommand{\topc}{\top}

\newcommand{\tuple}[1]{\langle #1 \rangle}

\newcommand{\wrt}{w.r.t.}

\title{Towards a  Forensic  Event  Ontology to Assist Video Surveillance-based Vandalism Detection}

\author[1]{Faranak Sobhani}  
\author[2]{Umberto Straccia} 
\affil[1]{Queen Mary University of London, UK}
\affil[2]{ISTI-CNR, Italy}

\begin{document}

\maketitle

\begin{abstract}
The detection and representation of events is a critical element in automated surveillance systems. We present here an ontology for representing complex semantic events to assist video surveillance-based vandalism detection. The ontology contains the definition of a rich and articulated event vocabulary  that is aimed at aiding forensic analysis to objectively identify and represent complex events.  Our  ontology has then been applied in the context of London Riots, which took place in 2011. We report also on the experiments conducted to support the classification of complex criminal events from video data.


\end{abstract}

\section{Introduction}
\label{sec:intro}

In the context of vandalism and terrorist activities, video surveillance forms an integral part of any incident investigation and, thus, 
%
%
there is a critical need for developing an ``automated video surveillance system" with the capability of detecting complex events to aid the forensic investigators in solving the criminal cases. 
As an example, in the aftermath of the London riots in August 2011 police had to scour through more than 200,000 hours of CCTV videos to identify suspects. Around 5,000 offenders were found by trawling through the footage, after a process that took more than five months. 

With the aim to develop an open and expandable video analysis framework equipped with tools for analysing, recognising, extracting and classifying events in video, which can be used for searching during investigations with unpredictable characteristics, or exploring normative (or abnormal) behaviours, several efforts for standardising event representation from surveillance footage have been made~\cite{francois2005verl,hakeem2005object,Hakeem04,nevatia2004ontology,nevatia2003hierarchical,scherp2009f,snidaro2007representing,westermann2007toward}.


While various approaches have relied on offering foundational support for the domain ontology extension, to the best of our knowledge, a systematic ontology for standardising the event vocabulary for forensic analysis and an application of it has not been presented in the literature so far.

In this paper, we present an OWL 2~\cite{OWL2} ontology for the semantic retrieval of complex events to aid video surveillance-based vandalism detection. Specifically, the ontology is a derivative of the DOLCE foundational ontology~\cite{sep-events} aimed to represent events that forensic analysts commonly encounter to aid in the investigation of criminal activities. The systematic categorisation of a large number of events aligned with the philosophical and linguistic theories enables the ontology for interoperability between surveillance systems. We also report on the experiments we conducted with the developed ontology to support the (semi-) automatic classification of complex criminal events from semantically annotated video data.

Our work significantly extends the preliminary works~\cite{Henderson15,Sobhani16}. The work~\cite{Henderson15} is an embryonal work investigating about the use of an ontology for automated visual surveillance systems, which then has been then further developed in~\cite{Sobhani16}. While our work shares with~\cite{Sobhani16} some basic principles in the development of the ontology, here  the level of details is now higher (\eg, the $\cp{Endurant}$ class (see Section~\ref{subsec:end} and its sub-classes have not been addressed in~\cite{Sobhani16}) and various ontological errors have been revised. Additionally, and somewhat more important, in our work experiments have been conducted for criminal event classification based on London 2011 riots videos.  Furthermore, but  less related, is~\cite{Sobhani15} in which the technical challenges facing researchers in developing computer vision techniques to process street-scene videos are addressed. The work focusses on standard image processing methods and does not deal with ontologies in any way.

The remainder of the paper is organised as follows. Related work is addressed  in Section~\ref{sec:litreview}. 
For the sake of completeness, Section~\ref{dls} presents complementary, theoretical material to ease the understanding of the ontology expressions used in our work.
Section~\ref{sec:for} presents a detailed description of the forensic ontology about complex criminal events. In Section~\ref{sec:aasist} we discuss how to use the ontology to assist video surveillance-based vandalism detection.
In Section~\ref{sec:validation} we conduct some experiments with our ontology based on CCTV footage of London riots from 2011, and finally,
 Section~\ref{sec:Conclusion} concludes.




\section{Related Work}
\label{sec:litreview}

In~\cite{nevatia2003hierarchical}, the \emph{Event Recognition Language} (ERL) is presented, which can describe hierarchical representation of complex spatiotemporal and logical events. The proposed event structure consists of primitive, single-thread, and multi thread events. Another event representation ontology, called CASE$^E$, is based on natural language representation and is proposed in~\cite{Hakeem04} and then extended in~\cite{hakeem2005object}. Subsequently, in~\cite{francois2005verl,nevatia2004ontology} a \emph{Video Event Representation Language} (VERL) was proposed for describing an ontology of events and the companion language called \emph{Video Event Markup Language} (VEML), which is a representation language for describing events in video sequences based on OWL~\cite{motik2009owl}. In~\cite{snidaro2007representing}, event detection is performed using a set of rules using the SWRL language~\cite{SWRL}.

The \emph{Event Model E}~\cite{westermann2007toward} has been developed based on an analysis and abstraction of events in various domains such as research publications, personal media~\cite{appan2004networked}, meetings~\cite{jain2003experiential}, enterprise collaboration~\cite{kim2004personal} and sports~\cite{pingali2002instantly}. The framework provides a generic structure for the definition of events and is extensible to the requirements ontology of events in the most different concrete applications and domains.

In~\cite{scherp2009f} a formal model of events is presented, called \emph{Event-Model-F}. The model is based on the foundational ontology DOLCE+DnS Ultralite (DUL) and provides comprehensive support to represent time and space, objects and persons, as well as mereological, casual, and correlative relationships between events. In addition, the Event-Model-F provides a flexible means for event composition, modelling event causality and event correlation, and representing different interpretations of the same event. 
The Event-Model-F is developed following the pattern-oriented approach of DUL, is modularised in different ontologies, and can be easily extended by domain specific ontologies. 

While the above-mentioned approaches essentially provide frameworks for the representation of events, none of them  address the problem of formalising forensic events in terms of a standard representation language such as OWL 2\footnote{We recall that the relationship to our  previous work~\cite{Henderson15,Sobhani16,Sobhani15} has been addressed in the introductory section} and, importantly, none have been applied and tested so far in a real use case,  which are the topics of the following sections. 

\section{Description Logics Basics} \label{dls}

For the sake of completeness of this work, we recap here succinctly some basic notions related to the \emph{Description Logics} (DLs) family of languages, which is the logical counterpart of the \emph{Ontology Web Language} (OWL 2)~\cite{OWL2}. We will  in fact use DL expressions to explain the ingredients of the forensic event ontology, which we will use in the following sections.
We  refer the reader to \eg~\cite{Baader03a,Baader09} for further insights on DLs.

Roughly, DLs allow  to describe classes (also called concepts), properties of classes and relationships among classes and properties.
Formally, we start with the basic DL called $\alc$ ($\mathcal{A}$ttributive $\mathcal{L}$anguage with $\mathcal{C}$omplement)~\cite{Schmidt-Schauss91}. 
Elementary descriptions are \emph{atomic concepts}, also called \emph{concept names}  (denoted $A$) and \emph{atomic roles} (denoted $r$). 
Complex \emph{concepts} (denoted $C$) can be built from them inductively with concept constructors. Specifically, concepts in $\alc$ are formed according to the following syntax rule:

\[
\begin{array}{llcll} 
C,D & \to & A & \mid & \mbox{(atomic concept)} \\
  &  & \topc & \mid & \mbox{(universal concept)} \\
  &  & \bottomc & \mid & \mbox{(bottom concept)} \\
  &  & C \andc D& \mid & \mbox{(concept conjunction)} \\
  &  & C \orc D& \mid & \mbox{(concept disjunction)} \\
  &  & \notc C& \mid & \mbox{(concept negation)} \\
  &  & \all r.C & \mid & \mbox{(universal restriction)} \\
  &  & \csome r.C&  & \mbox{(qualified existential restriction)} \ .
\end{array}
\]

\nd Sometimes, the abbreviation $\some R$ is used in place of the unqualified existential restriction $\csome R.\topc$. 

We further extend \alc~by allowing both (i) \emph{complex roles} (or,  simply \emph{roles}) defined inductively from atomic roles as follows ($r,s$ are roles):
\begin{itemize}
\item an atomic role is a role;
\item $r^-$ is a  role and denotes the inverse of role $r$;
\item $r \circ s$ is a role and denotes the composition of role $r$ and $s$;
\end{itemize}

\nd and (ii) \emph{existential value restrictions}, which are concepts of the form $\exists r.\oneof{a}$, where $a$ is an individual. 

We recall that, informally, a concept denotes a set of objects, while a role is a binary relation over objects.  So, for instance, the concept $\cp{Person} \andc  \some \role{hasChild}.\cp{Femal}$ will denote  people having a female child, while  $\cp{Person} \andc  \some \role{hasFriend}.\cp{\{mary\}}$ will denote  people having $\ind{mary}$ as friend.

A \emph{knowledge base} $\K$ consists of a finite set of axioms. An \emph{axiom} may be a

\begin{itemize}
\item  \emph{General Concept Inclusion axiom} (GCI) $C \impc D$, where $C$ and $D$ are concepts (read it as ``all instances of $C$ are instances of $D$''). 
An example of GCI is $\cp{Male} \impc \notc \cp{Female}$.

\item \emph{concept and role assertion axiom} $\cass{a}{C}$ and $\rass{a}{b}{r}$, respectively, where $a$ and $b$ are individuals. Examples of assertion axioms are $\cass{\ind{tim}}{\cp{Person}}$  (``tim is a person'') and  $\rass{\ind{tim}}{\ind{pat}}{\role{hasChild}}$ (``tim has pat as child'').

\item \emph{Role Inclusion axiom} (RI) $r \impc s$, where $r$ and $s$ are roles (read as ``all instances of role $r$ are instances of role $s$"). An example of RI is $\role{hasPart}\circ \role{hasPart}  \impc  \role{hasPart}$, which declares role $\role{hasPart}$ as transitive.

\end{itemize}

\nd From a semantics point of view, an {\em interpretation}  $\I$ is a pair $\I = (\highi{\Delta}, \highi{\cdot})$ consisting of a non-empty set $\highi{\Delta}$ (called the {\em domain}) and of an 
{\em interpretation function} $\highi{\cdot}$ that assigns to each atomic concept a subset of $\highi{\Delta}$, to each role a subset of $\highi{\Delta} \times \highi{\Delta}$ and to each individual $a$ an element in $\highi{\Delta}$.
%
%
Eventually, the mapping $\highi{\cdot}$ is extended to complex roles as follows:
\[
\begin{array}{rcl}
\highi{(r^-)} & = & \{ \tuple{x,y} \mid \tuple{y,x} \in \highi{r} \} \\
\highi{(r \circ s)} & = & \{ \tuple{x,y} \mid \exists z \mbox{ s.t. } \tuple{x,z} \in \highi{r} \mbox{ and }  \tuple{z,y} \in \highi{s} \} \ .
\end{array}
\]

\nd The mapping $\highi{\cdot}$ is extended to complex concepts as follows:
\[
\begin{array}{rcl}
\highi{\topc} & = & \highi{\Delta} \\ 
\highi{\bottomc}  & = &  \emptyset \\
\highi{(C \andc D)} & = & \highi{C} \cap \highi{D} \\
\highi{(C \orc D)} & = & \highi{C} \cup \highi{D} \\
\highi{(\notc C)} & = & \highi{\Delta} \setminus \highi{C} \\ 
\highi{(\all r.C)}  & = &  \{ x \in \highi{\Delta} \mid \highi{r}(x) \subseteq  \highi{C} \} \\
\highi{(\csome r.C)}  & = &  \{ x \in \highi{\Delta} \mid \highi{r}(x) \cap \highi{C} \neq \emptyset\} \\
\highi{(\csome r.\oneof{a})}  & = &  \{ x \in \highi{\Delta} \mid \highi{r}(x) \cap \{\highi{a}\} \neq \emptyset\} \ ,
\end{array}
\]

\nd where $\highi{r}(x) = \{y \colon \tuple{x,y} \in \highi{r}\}$. 

The \emph{satisfiability} of an axiom $E$ in an interpretation $\I = (\highi{\Delta}, \highi{\cdot})$, denoted $\I \models E$, is defined as  follows: 
$\I \models C \impc D$ iff $\highi{C} \subseteq \highi{D}$, 
$\I \models r \impc s$ iff $\highi{r} \subseteq \highi{s}$, 
$\I \models \cass{a}{C}$ iff $\highi{a} \in \highi{C}$, and
$\I \models \rass{a}{b}{r}$ iff $\tuple{\highi{a},\highi{b}} \in \highi{r}$.
Given a knowledge base $\K$, we say that $\I$ \emph{satisfies} $\K$ iff $\I$ satisfies each element in $\K$. If $\I \models \K$ we say that $\I$ is a \emph{model} of $\K$.  An axiom $E$ is a \emph{logical consequence}  of a knowledge base $\K$, denoted $\K \models E$, iff every model of $\K$ satisfies $E$. Determining whether $\KB \models \cass{a}{C}$ is called the \emph{instance checking problem}, while determining whether $\KB \models C \impc D$ is called the \emph{subsumption problem}.

\section{A Forensic Event Ontology}
\label{sec:for}

In the following, we present an OWL 2  ontology to support to some extent the semantic retrieval of complex events to aid automatic or semi-automatic video surveillance-based vandalism detection.  The idea is to develop an ontology that not only conveys a shared vocabulary, but some inferences based on it may assist a human being to support the video analysis by hinting to videos that may be more relevant than others in the detection of criminal events.

\subsection{The Role of a Foundation Ontology}
\label{sec:fouontology}

\begin{figure}[t!]
    \begin{center}
       \includegraphics[scale=0.4]{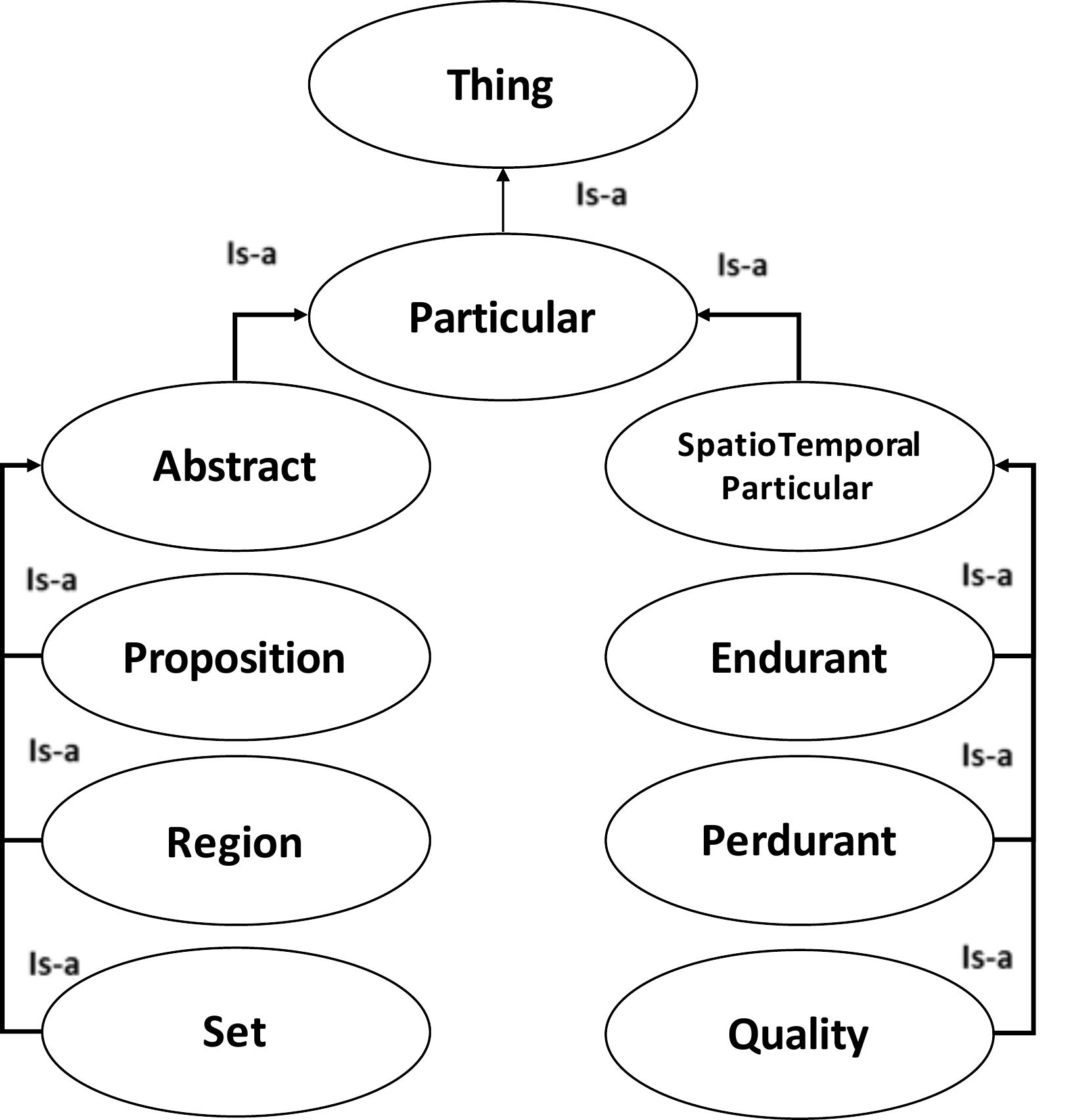}
        \caption{The foundational DOLCE ontology.}
        \label{fig:Onto}
    \end{center}
\end{figure}

To facilitate the elimination of the terminological ambiguity and the understanding and interoperability among people and machines~\cite{masolo2003wonderweb}, it is common practice to consider a so-called \emph{foundational ontology}. Let us note that several efforts have been taken by researchers in defining the foundational ontologies, such as BFO,\footnote{http://ifomis.uni-saarland.de/bfo/} SUMO,\footnote{http://www.adampease.org/OP/} UFO\footnote{https://oxygen.informatik.tu-cottbus.de/drupal7/ufo/} and DOLCE,\footnote{http://www.loa.istc.cnr.it/old/Papers/DOLCE2.1-FOL.pdf} to name a few. AS DOLCE ontology offers a cognitive bias with the ontological categories underlying natural language and human common sense, the same is selected for our proposed extension.  We recall that the DOLCE foundational ontology (see Figure~\ref{fig:Onto}) encompasses \onto{Endurant} and \onto{Perdurant} entities. \onto{Endurant} entities are ever-present at any time as opposed to \onto{Perdurant} entities that extended in time by accumulating different temporal parts. A more thorough explanation on the DOLCE events conceptualisation can be found \eg~in~\cite{sep-events}.

\subsection{A Forensic Complex Event Ontology}
\label{sec:dolce}    

Our complex event classes extend DOLCE's \onto{Perdurant} class. To assign the action classes into respective categories, we follow a four-way classification of action-verbs: namely, into \onto{State}, \onto{Process}, \onto{Achievement} and \onto{Accomplishment} using event properties such as \onto{telic}, \onto{stage} and \onto{cumulative} (see~\cite{rothstein2004verb,vendler1957verbs,vendler1967linguistics}). The distinction between these concepts are derived from the event properties as illustrated in Table~\ref{tab:event}, which we summarise below.

\begin{table}
    \centering
    \caption{Classification of Event Types.}
    \label{tab:event}
    \begin{tabular}{|l|l|l|l|}
        \hline
        \onto{State} & -telic & -stage & cumulative  \\ \hline
        \onto{Process} & -telic & +stage & -  \\ \hline
        \onto{Achievement} & +telic & -stage & not cumulative \\ \hline
        \onto{Accomplishment} & +telic & +stage & not cumulative \\ \hline    
    \end{tabular}
\end{table}

\begin{itemize}
\item \textbf{State [-telic,-stage]} This action category represents a long, non-dynamic event in which every instance is the same: there cannot be any distinction made between the stages. States are cumulative and homogenous in nature.

\item \textbf{Process [-telic, +stage]} The action category, like \onto{State}, is atelic, but unlike  \onto{State}, the action undertaken are dynamic. The actions appear progressively and thus can be split into a set of stages for analysis.

\item \textbf{Accomplishment [+telic, +stage]} Accomplishments are telic and cumulative activities, and thus behave differently from both \onto{State} and \onto{Process}. The performed action can be analysed in stages and in this way, they are similar to \onto{Process}. Intuitively, an accomplishment is an activity which moves toward a finishing point as it has variously been called in the literature. Accomplishment is also cumulative activity.

\item \textbf{Achievement [+telic, -stage]} Achievements are similar to \onto{Accomplishment} in their telicity. They are also not cumulative with respect to contiguous events.
Achievements do not go on or progress, because they are near instantaneous, and are over as soon as they have begun.

\end{itemize}

\subsubsection{Forensic Perdurant Entities}
\label{subsec:dolce}

Perdurant entities,  extend in time by accumulating different temporal parts and some of their proper temporal parts may be not present. To this end,  \onto{Perdurant} entities are divided into the classes \onto{Event} and \onto{Stative}, classified according to their temporal characteristics. 

\begin{figure}
\begin{center}
\includegraphics[scale=0.6]{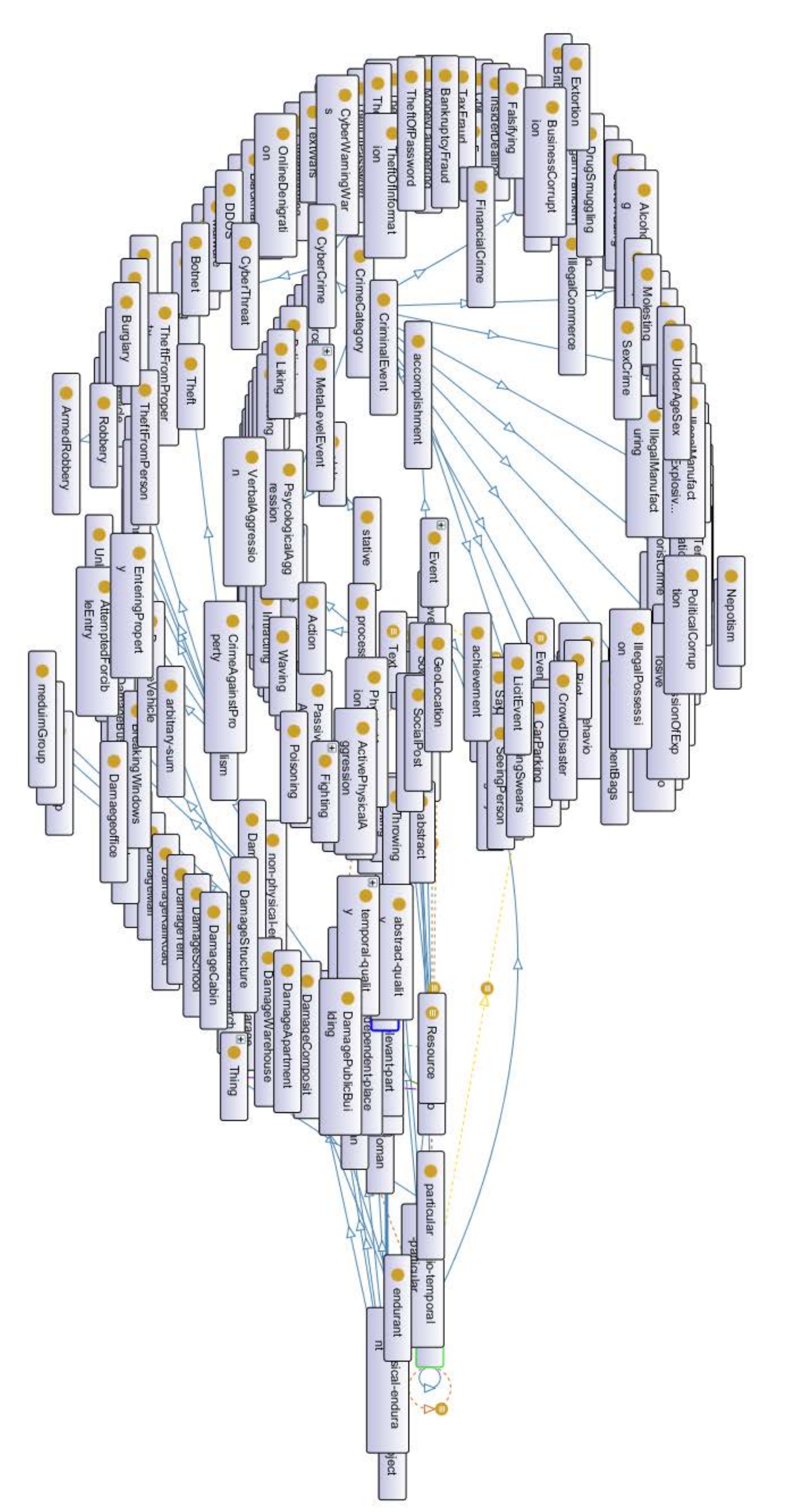}
\caption{The forensic extension of the $\cp{Perdurant}$  class.}\label{fig:Perdurant}
\end{center}
\end{figure}

Axiom sets (\ref{ax8})-(\ref{ax4}) below provide a subset of our formal extension of the \onto{Perdurant}  vocabulary.
The forensic extension of the ontology structure is shown in Figure~\ref{fig:Perdurant}.
 
\begin{equation}\label{ax8}
\begin{array}{rcl}
\cp{Perdurant} & \sqsubseteq & \cp{SpatioTemporalParticular} \\
\cp{Perdurant} &\sqsubseteq & \exists \role{participant}.\cp{Endurant}\\
\cp{Fighting} &\sqsubseteq &\exists \role{participant}.\cp{GroupOfPeople} \\
\cp{Perdurant} & \sqsubseteq &\neg  \cp{Endurant} \\
\cp{Kicking} &\sqsubseteq &\neg \cp{Vehicle}
\end{array}
\end{equation}
    
\begin{equation}\label{ax7}
\begin{array}{rcl}
\cp{State} & \sqsubseteq & \cp{Stative} \\
\cp{MetaLevelEvent} &\sqsubseteq & \cp{State}\\
\cp{Accusing} &\sqsubseteq &\cp{MetaLevelEvent} \\
\cp{Beiliving} & \sqsubseteq &\cp{MetaLevelEvent} \\
\cp{PsycologicalAggression} &\sqsubseteq &\cp{State}\\
\cp{Blaming} & \sqsubseteq &\cp{PsycologicalAggression} \\
\cp{Bullying} &\sqsubseteq &\cp{PsycologicalAggression}\\
\end{array}
\end{equation}
 
\begin{equation} \label{ax6}
\begin{array}{rcl}
\cp{Process} & \sqsubseteq & \cp{Stative} \\
\cp{Action} &\sqsubseteq & \cp{Process}\\
\cp{Gesture} &\sqsubseteq &\cp{Process} \\
\cp{PhysicalAggresion} & \sqsubseteq &\cp{Process} \\
\cp{ActivePhysicalAggresion} &\sqsubseteq &\cp    {PhysicalAggresion}\\
\end{array}
\end{equation}

\begin{equation}\label{ax5}
\begin{array}{rcl}
\cp{Accomplishment} & \sqsubseteq & \cp{Event} \\
\cp{CriminalEvent} &\sqsubseteq & \cp{Accomplishment}\\
\cp{EventCategory} &\sqsubseteq &\cp{Accomplishment} \\
\cp{Crimecategory} & \sqsubseteq &\cp{Stative} \\
\end{array}
\end{equation}

\begin{equation} \label{ax4}
\begin{array}{rcl}
\cp{Achievement} & \sqsubseteq & \cp{Event} \\
\cp{Saying} &\sqsubseteq & \cp{Achievement}\\
\cp{Seeing} &\sqsubseteq &\cp{Achievement} \ .
\end{array}
\end{equation}

\begin{figure}[t!]
    \begin{center}
		\includegraphics[scale=0.4]{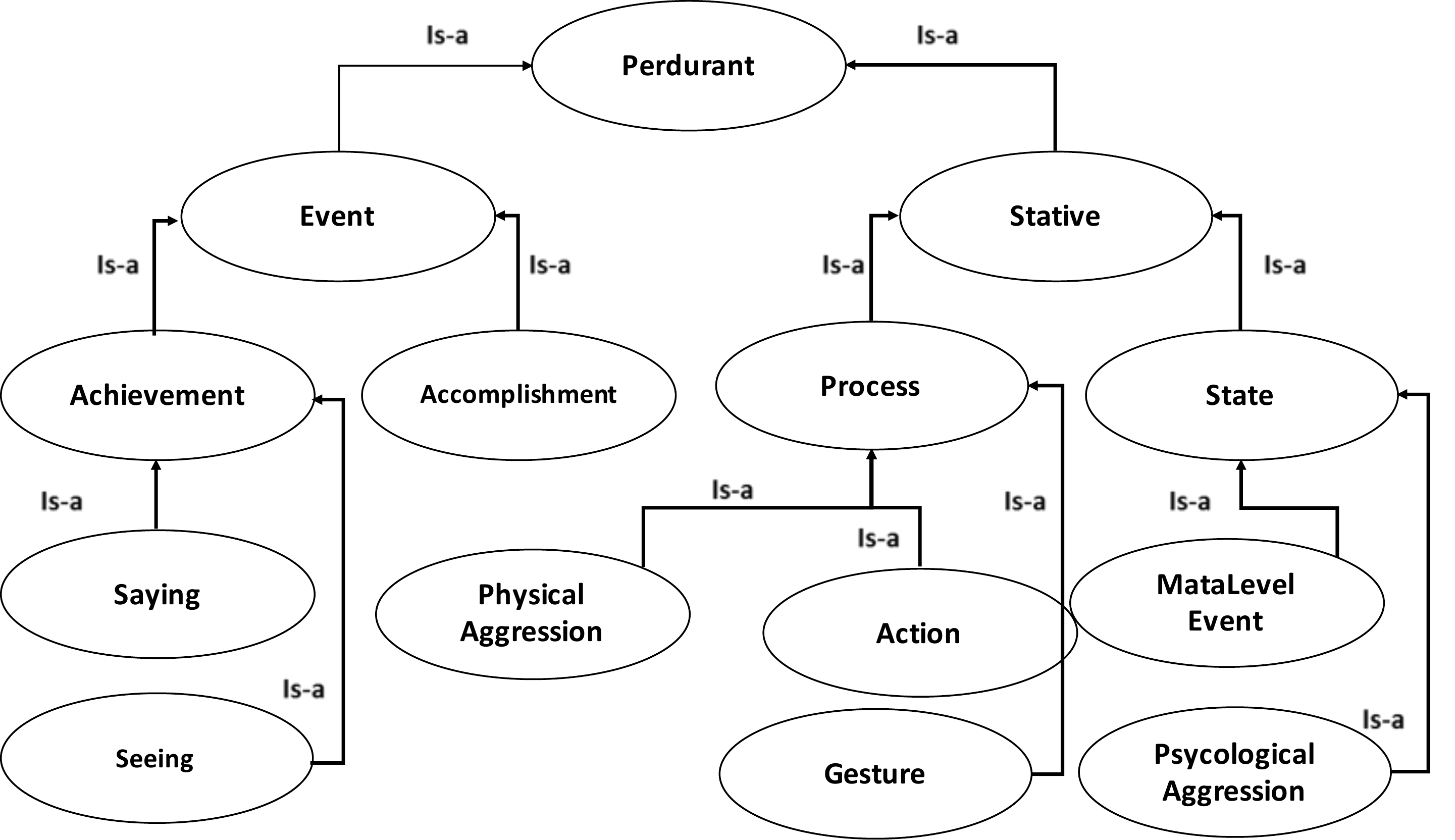}
        \caption{The $\cp{Perdurant}$ class hierarchy  for forensic events descriptions.}
        \label{fig:dolceA}
    \end{center}
\end{figure}

\nd An excerpt of the forensic ontology is shown in Figure~\ref{fig:dolceA} and strictly adheres to the above terminological determination of action categories as mentioned before and extends the classes with suitable event concepts. 

The concept $\cp{State}$ offers representation for  $\cp{MetaLevelEvent}$ which encompasses abstract human events such as $\cp{Accusing}$, $\cp{Believing}$ and $\cp{Liking}$ among others. As previously stated, the concept $\cp{State}$ represents a collection of events which are exhibited by a human that is time-consuming, non-dynamic, cumulative and homogenous. The other sub-class of $\cp{State}$ is $\cp{PsychologicalAggression}$ which characterises the human actions such as $\cp{Blaming}$, $\cp{Decrying}$, $\cp{Harassing}$ and so forth.
The concept $\cp{Process}$ includes several human action categories that represent dynamic events which can be split into several intermediate stages for analysis. For the purposes of clarity, the concept $\cp{Process}$ offers three sub-concepts namely $\cp{Action}$, $\cp{Gesture}$ and $\cp{PhysicalAggression}$. The $\cp{Action}$ class incorporates different event such as $\cp{Dancing}$, $\cp{Greeting}$, $\cp{Hugging}$ among other concepts defined. The concept $\cp{Gesture}$ formalises the different interest points related to human gestures. In order to eliminate the ambiguity traditionally present in human gestures across cross-cultural impact, the action performed during the gesture is captured and represented in the ontology and, thus, enabling the removal of subjectivity from the concept definition. The final sub-class of the $\cp{Process}$ class includes the concept $\cp{PhysicalAggression}$ and formalises human conflicting actions.

By and large, the human action categorised into \onto{State} and \onto{Process} represent the microscopic movements of humans. 

\begin{figure}[t!]
    \begin{center}
           \includegraphics[scale=0.4]{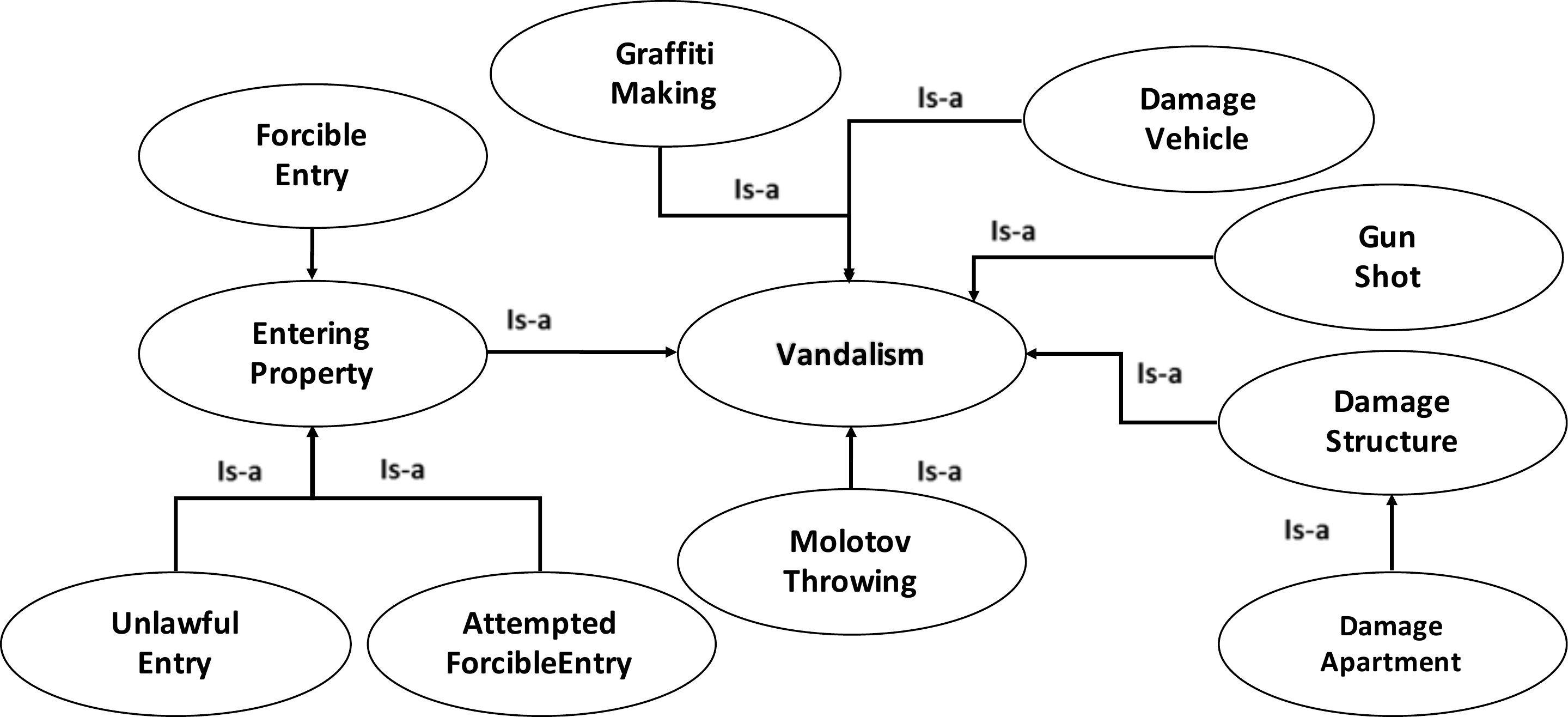}
        \caption{The concept hierarchy of $\cp{Vandalism}$, direct subclass of $\cp{CrimeAgainstProperty}$. The latter is a subclass of class $\cp{Accomplishment}$.}
        \label{fig:dolceB}
    \end{center}
\end{figure}

\begin{figure}[t!]
    \begin{center}
           \includegraphics[scale=0.4]{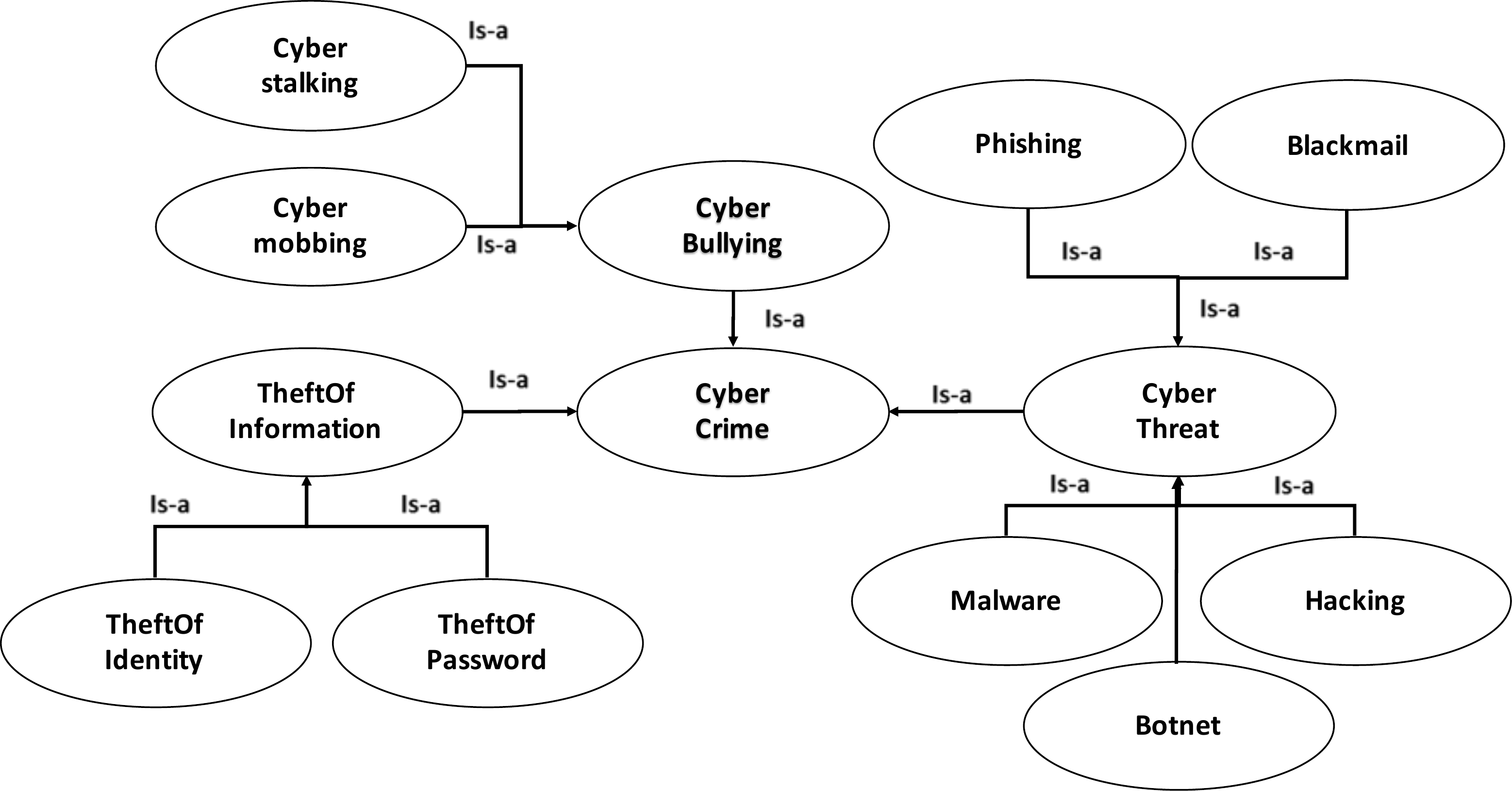}
        \caption{The concept hierarchy of $\cp{CyberCrime}$.}
        \label{fig:dolce2}
    \end{center}
\end{figure}

From the automatic surveillance viewpoint, these microscopic events may be extracted from media items. In contrast, the event representation formalised by means of the concepts $\cp{Achievement}$ and $\cp{Accomplishment}$ offer a rich combination of human events that allow for the construction of complex events with or without the combination of microscopic features. For instance, the concept hierarchy for \onto{Vandalism} is illustrated in Figure~\ref{fig:dolceB}, while the concept hierarchy for \onto{CyberCrime} is shown in Figure~\ref{fig:dolce2} instead.

\subsubsection{Forensic Endurant Entities}
\label{subsec:end}

DOLCE is based on fundamental distinction among $\cp{Endurant}$ and $\cp{Perdurant}$ entities. The difference between $\cp{Endurant}$ and $\cp{Perdurant}$ entities is related to their behaviour in time. Endurant are wholly present at any time they are present. Philosophers believe that endurant are entities that are in time while lacking, however, temporal parts~\cite{masolo2003wonderweb}. Therefore, the proposed vocabulary structure of all possible forensic entities also extends on $\cp{Endurants}$ entities.

\begin{figure}[t!]
    \begin{center}
       \includegraphics[scale=0.5]{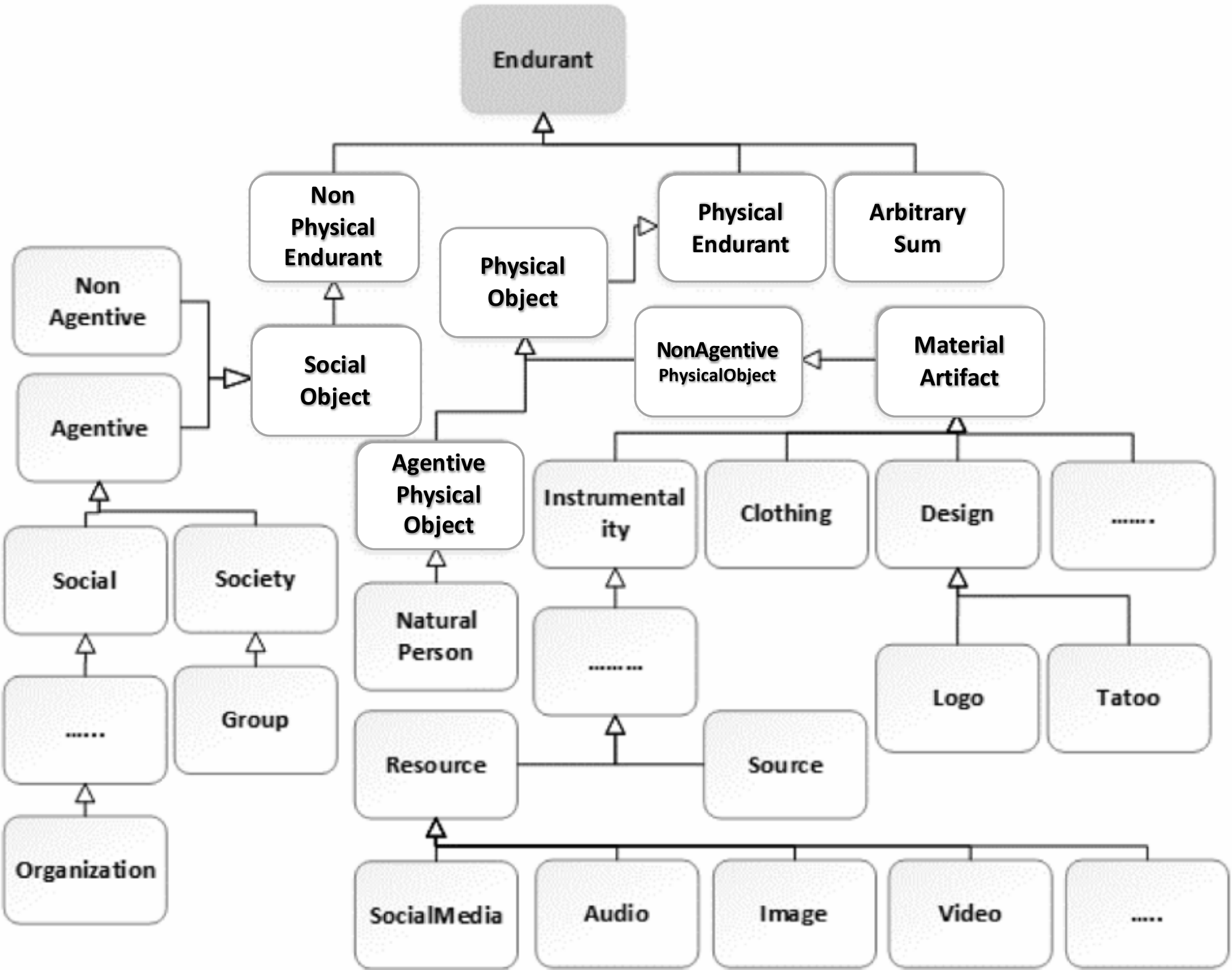}
        \caption{Excerpt of the $\cp{Endurant}$ concept hierarchy in the forensic ontology.}
        \label{fig:Endurant}
    \end{center}
\end{figure}

Axiom set~(\ref{ax1}) describes a subset formalization of the $\cp{Endurant}$  vocabulary 
and an excerpt of the forensic extension of the ontology structure shown in Figure~\ref {fig:Endurant}.

\begin{equation}  \label{ax1}
\begin{array}{rcl}
\cp{Endurant} & \sqsubseteq & \cp{SpatioTemporalParticular} \\
\cp{Endurant} &\sqsubseteq & \exists \role{participantIn}.\cp{Perdurant}\\
\role{participantIn} &= &  \role{participant}^-\\
\cp{NonPhysicalEndurant} &\sqsubseteq & \cp{Endurant}\\
\cp{PhysicalEndurant} &\sqsubseteq & \cp{Endurant}\\
\cp{ArbitrarySum} &\sqsubseteq  & \cp{Endurant} \ .
\end{array}
\end{equation}

\section{Assisting Video Surveillance-based Vandalism Detection}
\label{sec:aasist}

We next show how the so far developed ontology is expected to be used to assist video surveillance-based vandalism detection.

\subsection{Annotating Media Objects, viz. Surveillance  Videos}
\label{sec:datas}

Given surveillance videos and any media in general, we need a method to annotate them by using the terminology provided by our ontology.
This gives rise to a set of facts that, together with the inferred facts, may support a more effective automatic or, more likely, semi-automatic retrieval of relevant information, such as \eg~vandalic acts.
Specifically, the inferred information may suggest a user look at some \eg~video sequences or video still images, rather than to others first.

\begin{figure}
    \begin{center}
                \includegraphics[scale=0.45]{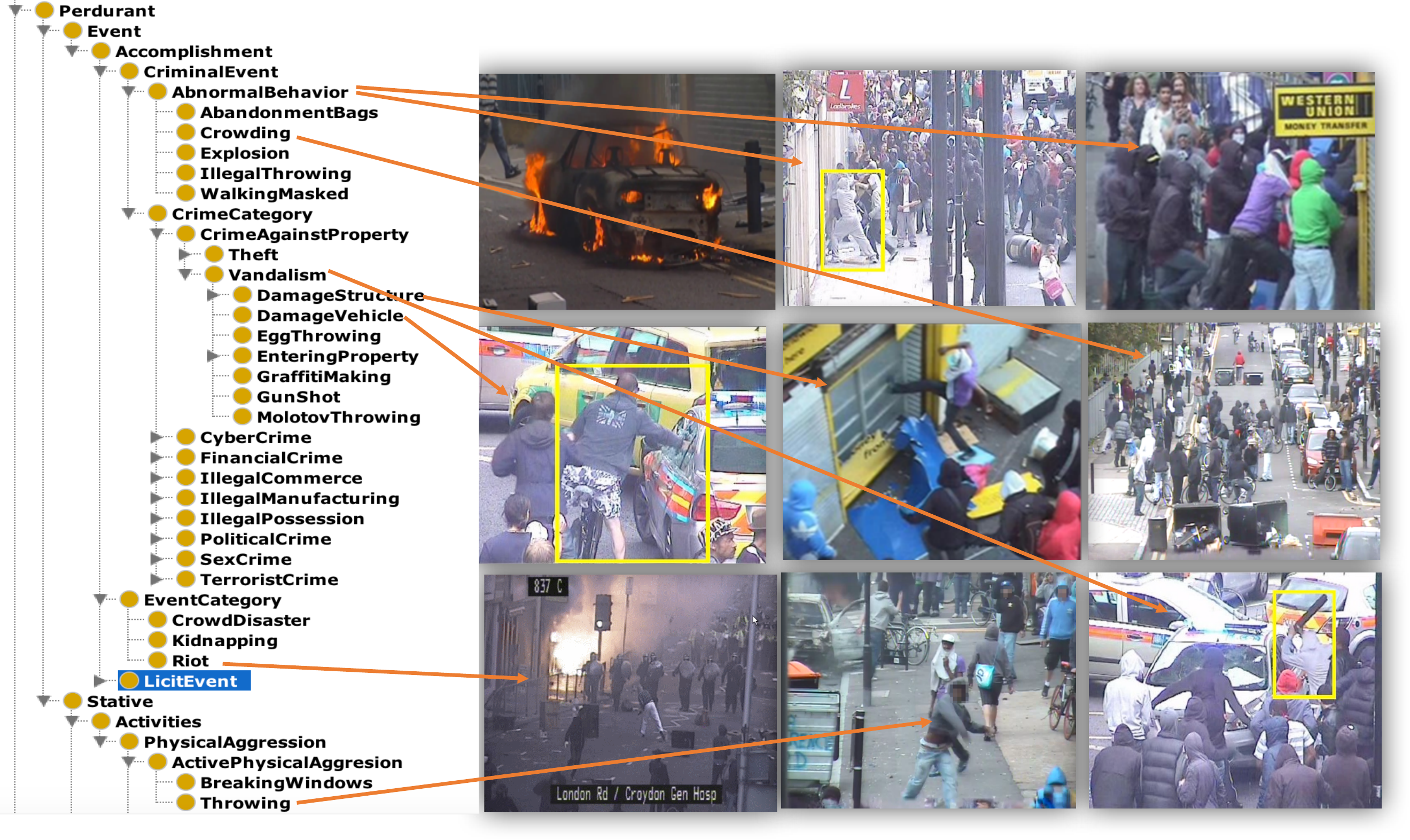}
        \caption{Examples of still image annotations from the London Riots 2011 of events as per Table~\ref{crimevents}.}
        \label{annot}
    \end{center}
\end{figure}

The general model we are inspired on is based on~\cite{Meghini01}. Conceptually, according to~\cite{Meghini01}, a media object $o$ (\eg, an image region, a video sequence,  a piece of text, etc.) is annotated with one (or more) entities $t$ of the ontology (see \eg~Figure~\ref{annot}).

For instance, stating that an image object $o$ \emph{is about} a $\cp{DamageVehicle}$ can be represented conceptually via the DL expression 
\[
\cass{o}{(\some \role{isAbout}.\cp{DamageVehicle})} \ .
\]
\nd As specified in~\cite{Meghini01}, such an annotation may come manually from a user or, if, available, from an image classifier. In the latter case, it may annotate the image automatically, or, semi-automatically by suggesting to a human \emph{annotator}, which are the most relevant entities of the ontology that may be used for a specific media object $o$.
%
%
%
Note, however, that, the above methodology just illustrates the concept. In our case, for the sake of ease the annotation,
we may not enforce the use of the object property $\role{isAbout}$ (see Example~\ref{ex2} later on). 
Generally, we will annotate a $\cp{Resource}$ with  $\cp{Perdurant}$s and $\cp{Endurant}$s: thus, if an image is annotated with \eg~a perdurant that is a damaged vehicle, then this means that the image \emph{is about} a damaged vehicle.

\begin{figure}
    \begin{center}
                \includegraphics[scale=0.3]{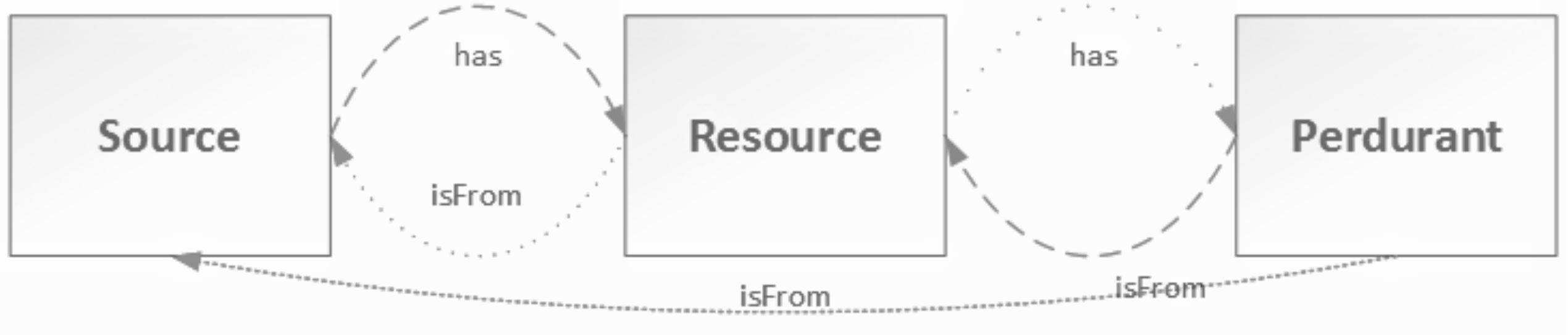}
        \caption{Modelling transitive property axiom in our defined ontology.}
        \label{fig:inverse}
    \end{center}
\end{figure}

 We recall that $\cp{Resource}$s (and $\cp{Source}$s) are modelled as follows (see Figure~\ref{fig:inverse}):
\[
\begin{array}{lcl}
\cp{Source} & \impc  & \cp{Endurant} \andc \some \role{has}.\cp{Resource}  \\
&& \andc \some \role{hasCameraId}.\cp{string} \\
&& \andc \some \role{hasLatitude}.\cp{string} \\
&& \andc \some \role{hasLongitute}.\cp{string} \\
&& \andc \some \role{hasLocationName}.\cp{string} \\
\cp{Resource} & \impc & \cp{Endurant} \andc \some \role{has}.\cp{Perdurant} \\
\role{has}  & = & \role{isFrom}^- \\
\role{has} \compr \role{has} & \impc & \role{has} \ .
\end{array}
\]
\nd Note that in the last RI  axiom, $\compr$ is role composition and, thus, $\role{has} \compr \role{has}  \impc  \role{has}$ dictates that the property $\role{has}$ is transitive, while
with $\role{has}   =  \role{isFrom}^-$ we say that $\role{isFrom}$ is the inverse of $\role{has}$. Therefore, $\role{isFrom}$ is transitive as well. 

The following example illustrates the mechanism of image of annotation together with a meaningful inference.

\begin{example} \label{ex}
Consider the following DL axioms resulting from annotating images of a video ($\ind{video6}$) registered by a camera ($\ind{cameraC004}$):
\[
\begin{array}{l}
\rass{\ind{personA}}{\ind{throwing5}}{\role{participateIn}} \\
\cass{\ind{throwing5}}{\cp{Throwing}} \\
\cass{\ind{personA}}{\cp{NaturalPerson}} \\
\cp{Throwing} \impc \cp{ActivePhysicalAggression} \\
\cp{ActivePhysicalAggression} \impc \cp{PhysicalAggression} \\
\cp{PhysicalAggression} \impc \cp{Process} \\
\rass{\ind{throwing5}}{\ind{endurant6}}{\role{isFrom}} \\
\cass{\ind{endurant6}}{\cp{Resource}} \\
\rass{\ind{endurant6}}{\ind{video6}}{\role{hasVideoId}} \\
\cass{\ind{endurant7}}{\cp{Source}} \\
\rass{\ind{endurant7}}{\ind{cameraC004}}{\role{hasCameraId}} \\
\rass{\ind{endurant7}}{\ind{endurant6}}{\role{has}} \ .
\end{array}
\]
\nd Now, as $\role{isFrom}$ is transitive, we may infer: 
\[
\rass{\ind{throwing5}}{\ind{endurant7}}{\role{isFrom}} \ .
\]
\nd Then, it is not difficult to see that we  finally infer
\[
\cass{\ind{PersonA}}{
\begin{array}[b]{l}
\some \role{paticipateIn}.(\cp{PhysicalAggression} \ \andc\\
\hspace*{2.1cm} \some \role{isFrom}.(\cp{Source} \ \andc \
\some \role{hasCameraID}.\{\ind{cameraC004}\})) 
\end{array}
} \ ,
\]
\nd which can be read as:
\begin{quote}
``A person ($\ind{PersonA}$) participated in a physical aggression  that has been registered by camera C004".
\end{quote}
\qed
\end{example}

\subsection{Modelling GCIs for Vandalism Event Detection} 
\label{secrules}

As we are focusing on forensic domain and dealing with variety of concepts aiming at aiding forensic analysis, to objectively identify and represent  complex events, 
we next show that  a (manually build) \emph{General Concept Inclusion} (GCI) axiom may help to classify high-level events in terms of a composition of some lower level events.
The following are such GCI examples:

\begin{figure}
    \begin{center}
                \includegraphics[scale=0.75]{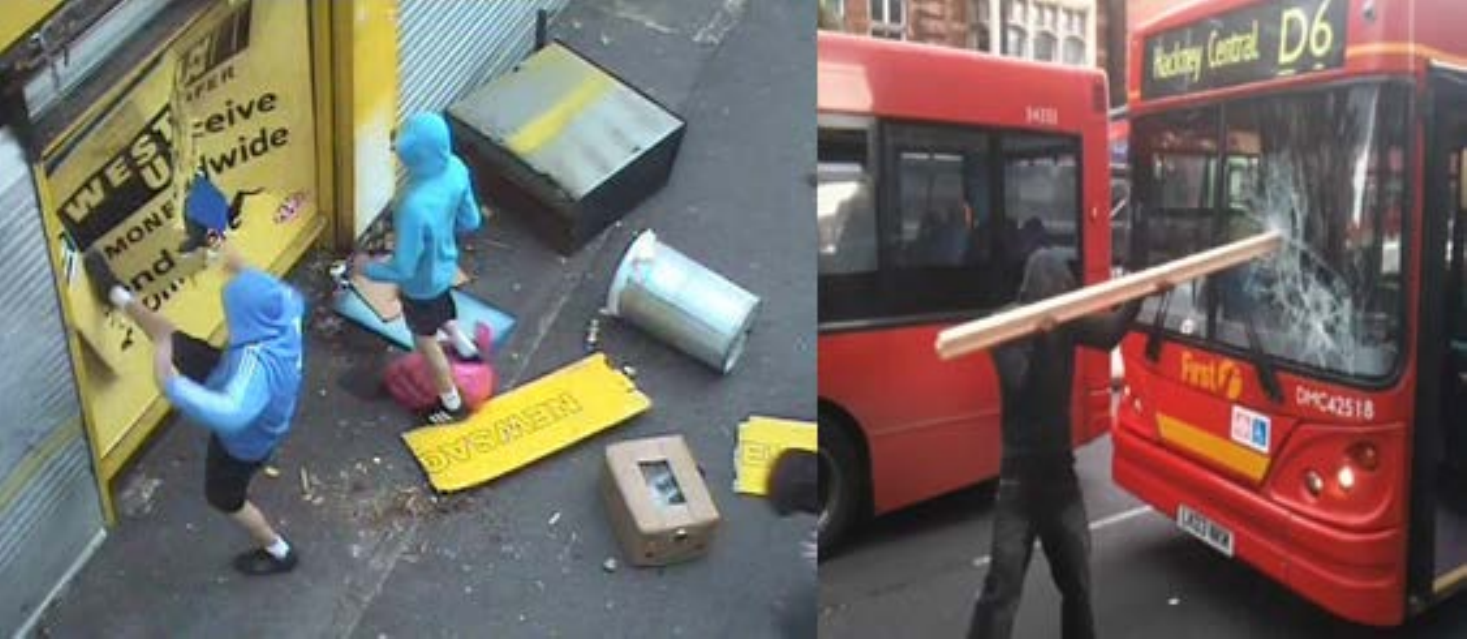}
        \caption{Example of  $\role{DamageVehicle}$ and  $\role{DamageStructure}$  scenes in  CCTV.}\label{fig:DV}
    \end{center}
\end{figure}

\begin{description}
\item[DamageVehicle:]
\[
\begin{array}{l}
\cp{Perdurant}\  \sqcap \\
\  \exists \role{participant}. (\cp{Vehicle} \ \sqcap \\ 
\hspace*{16ex} \exists \role {participantIn}. (\cp{BreakingDoor} \ \sqcup \  \cp{BreakingWindows}))\\
\hspace*{8cm}  \impc   \cp{DamageVehicle} \ .
\end{array}
\]

\begin{quote}
\emph{``If an event involves a vehicle that is subject of a breaking door or breaking windows then the event is about a damaged vehicle"} (see Figure~\ref{fig:DV}).
\end{quote}
\vspace*{3ex}

\item[DamageStructure:]

\[
\begin{array}{l}
\cp{Perdurant}  \ \sqcap \\
\ \exists \role{participant}.(\cp{Structure}  \ \sqcap \\
\hspace*{16ex}   \exists {\role{participantIin}}. \cp{Kicking})  \impc \cp{DamageStructure} \ .
\end{array}
\]

\begin{quote}
\emph{``If an event involves a structure that is subject of kicking,  then the event is about a damaged structure"} (see Figure~\ref{fig:DV}).
\end{quote}

\end{description}

\nd The following example illustrates the use of such GCIs.

\begin{example} \label{ex2}
 Suppose we have an image classifier that is able to provide us with the following facts. Specifically, assume it is able to identify vehicles and breaking windows:
 \[
\begin{array}{l}
\rass{\ind{Perdurant2}}{\ind{Endurant1}}{\role{participant}} ,  \cass{\ind{Endurant1}}{\cp{Vehicle}} \\
\cass{\ind{Perdurant2}}{\cp{BreakingWindows}} \ .
\end{array}
\]
 
\nd From these facts and the GCI about $\cp{DamageVehicle}$,  we may infer that the image is about a damaged vehicle, \ie~we may infer
\[
\cass{\ind{Perdurant2}}{\cp{DamageVehicle} } \ .
\] 
\qed
\end{example}

\begin{figure}
    \begin{center}
                \includegraphics[scale=0.4]{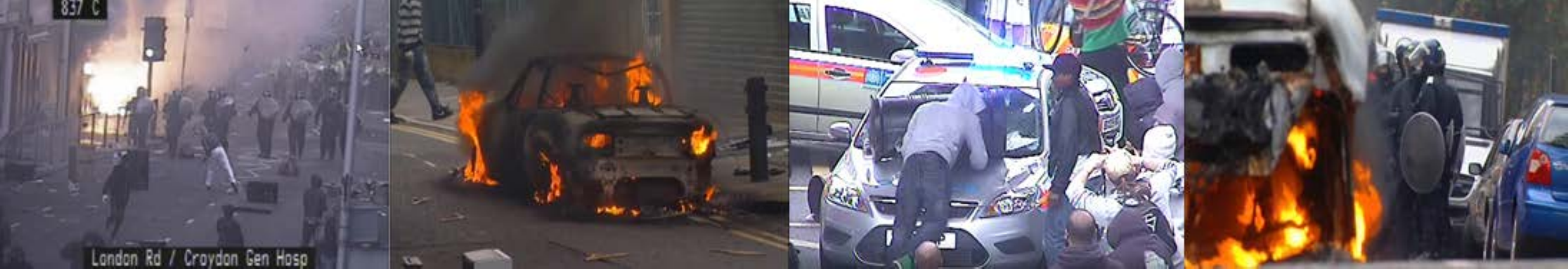}
        \caption{Example of $\cp{Vandalism}$ scenes in CCTV videos.}\label{fig:vandali}
    \end{center}
\end{figure}

\nd The following set of GCIs illustrates instead how one may have multiple GCIs to classify a single event, such as those for $\cp{Vandalism}$ (see, \eg~Figure~\ref{fig:vandali}).\footnote{Recall that all these GCIs provide sufficient conditions to be an instance of $\cp{Vandalism}$, but no necessary condition.} 

 \[
    \begin{array}{l}
    \cp{Perdurant} \sqcap \exists \role{part}.  (\cp{Crowding} \sqcap \cp{DamageStructure}) \sqsubseteq \cp{Vandalism}\\
    \cp{Perdurant} \sqcap  \exists \role{part} . (\cp{Crowding} \sqcap \cp{DamageVehicle}) \sqsubseteq \cp{Vandalism}\\
    \cp{Perdurant}  \sqcap \exists \role{part} . (\cp{Explosion}  \sqcap \cp{Throwing}) \sqsubseteq \cp{Vandalism} \ .
    \end{array}
    \]

\nd Note that in the example above, we assume that events (perdurant) may be complex in the sense that they may compose by multiple sub-events (parts). So, \eg~in the last GCI, we roughly state

\begin{quote}
\emph{``If a (complex) event involves both throwing and an explosion (two sub-events) then the event is about vandalism"}.
\end{quote}

\nd Following our previous examples, we next are going to formulate another kind of background knowledge. Our main focus in this example is on recognizing high-level events,
which occur in the same location (same street in our modelling). In order to model this scenario, we may use the \emph{Semantic Web Rule Language} (SWRL) to model the 
$\role{locatedSameAs}$ role and then use it in GCIs. The SWRL rule is:
\begin{quote}
\emph{``Two perdurants that occur in the same street occur in the same place.}
\end{quote}

\[
\begin{array}{l}
\cp{Perdurant}(?p1),\cp{Perdurant}(?p2), 
\role{hasLocationName}(?p1, ?l1), \\
\hspace*{4ex} \role{hasLocationName}(?p2, ?l2),
\role{SameAs}(?l1, ?l2) \rightarrow \role{locatedSameAs}(?p1, ?p2) \ .
\end{array}
\]

\begin{figure}
    \begin{center}
                \includegraphics[scale=0.4]{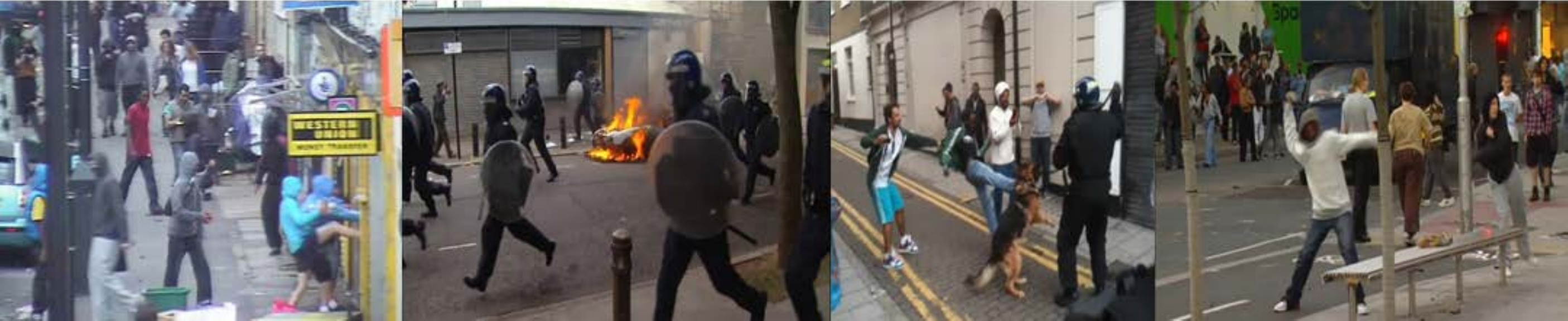}
        \caption{Examples of events that happen in the same location ($\role{locatedSameAs}$) from CCTV.}\label{fig:SL}
    \end{center}
\end{figure}

\nd The following axioms illustrate how to use the previously defined relation (few examples captured from our data set by these rules are illustrated in Figure~\ref{fig:SL}).
\[
\begin{array}{l}
    \cp{Perdurant} \  \sqcap \\
\ \    \exists \role{part} .(\cp{Crowding} \ \sqcap  \exists \role{locatedSameAs} . \cp{Explosion})   \sqsubseteq \cp{Vandalism} \\ \\
    \cp{Perdurant} \ \sqcap \\
  \  \ \exists\role{part}. (\cp{Crowding} \ \sqcap    \exists \role{locatedSameAs}. \cp{DamageStructure}) \sqsubseteq \cp{Vandalism} \\ \\
    \cp{Perdurant}   \ \sqcap \\
\  \  \exists \role{part}.(\cp{Crowding} \ \sqcap \exists \role{locatedSameAs} .\cp{Throwing}) \sqsubseteq \cp{Vandalism} \\\\
    \cp{Perdurant} \ \sqcap \\
\  \   \exists \role{part} . (\cp{DamageStructure} \ \sqcap  \exists \role{locatedSameAs} . \cp{Throwing}) \sqsubseteq \cp{Vandalism} \ .
\end{array}
\]    

\section{Experiments}
\label{sec:validation}

We conducted two experiments with our ontology, which we are going to describe in the following.\footnote{The ontologies used in the experiments and experimental results can be found at \url{http://www.umbertostraccia.it/cs/ftp/ForensicOntology.zip}.}

In the first case, we evaluated the classification effectiveness of manually built GCIs to identify crime events, while in the second case we drop the manual-built GCIs and, try to learn such GCIs instead automatically from examples and compare their effectiveness with respect the manually built ones.

\subsection{Classification via Manually Built GCIs} \label{valgci}

Roughly, we have considered a number of crime videos, annotated them manually and then checked whether the manually built GCIs, as described in Section~\ref{secrules}, were able to determine crime events correctly.

\vspace*{1ex}
\nd {\bf Setup.} Specifically, we considered our ontology and around  3.07 TB of video data about the London riot 2011,\footnote{These are part of the EU funded project LASIE ``Large Scale Information Exploitation of Forensic Data", \url{http://www.lasie-project.eu}.} of which 929 (GB) is in a non-proprietary format.
We considered 140 videos (however, the videos cannot be made publicly available). Within these videos,  all the available CCTV cameras (35 CCTV) along with their features such as latitude, longitude, start time, end time and street name, have been annotated manually according to our methodology described in Section~\ref{sec:aasist} and included into our ontology. We have also calculated all the geographic distances between each camera. The resulting ontology contains 1800 created individuals of which, \eg~106 are of type $\cp{Event}$.
\begin{table}
    \caption{Criminal event classes considered.} \label{crimevents}
    \[
    \begin{array}{|lll|} \hline
    \cp{Vandalism}  \ (13, 57) & \cp{Riot} \  (4, 21) & \cp{AbnormalBehavior} \ (2, 80) \\
      \cp{Crowding}  \ (1, 64) &   \cp{DamageStructure} \ (3, 9) &  \cp{DamageVehicle}  \ (3, 16) \\
      \cp{Throwing}  \ (1,30) & & \\ \hline
    \end{array}
    \]
\end{table}


\begin{table*}
\caption{Ontology Metrics.}\label{tab:metric}
\begin{center}
{\footnotesize
\begin{tabular}{cc}
\begin{tabular}{|l|l|}            \hline
Axioms & 9889    \\ \hline
Logical axiom count & 7176    \\ \hline
Class count     & 483 \\ \hline
Object property count & 148\\ \hline
Data property count & 51  \\ \hline    
Individual  count & 1800 \\ \hline    
DL expressivity & SHIQ(D) \\ \hline    
\end{tabular}
&        
\begin{tabular}{|l|l|}    \hline
SubclassOf axioms count & 532   \\ \hline
EquivalentClasses axioms count     & 5  \\ \hline
DisjointClasses axioms count & 11 \\ \hline
GCI count & 38  \\ \hline    
Hidden GCI Count  & 5 \\ \hline    
\end{tabular}
\\ \\
\begin{tabular}{|l|l|}                    \hline
SubObjectPropertyOf axioms count & 93  \\ \hline
InverseObjectProperties axioms count & 20  \\ \hline
TransitiveObjectProperty axioms count & 5 \\ \hline
SymmetricObjectProperty axioms count & 2\\ \hline    
ObjectPropertyDomain axioms count & 19 \\ \hline
ObjectPropertyRange axioms count & 18 \\ \hline    
\end{tabular}
&
\begin{tabular}{|l|l|}        \hline
SubDataProperty axioms count & 11  \\ \hline
DataPropertyDomain axioms count & 1 \\ \hline
DataPropertyRange axioms count & 5 \\ \hline    
\end{tabular} \\
\\ \\
\begin{tabular}{|l|l|}    \hline
ClassAssertion axioms count & 1793  \\ \hline
ObjectPropertyAssertion axioms count & 2964 \\ \hline
DataPropertyAssertion axioms count & 1706 \\ \hline    
\end{tabular}
&
\begin{tabular}{|l|l|}    \hline
AnnotationAssertion axioms count & 195  \\ \hline
\end{tabular} 
\end{tabular}
}
\end{center}
\end{table*}

Then, we considered criminal events occurring in the videos (specifically, we focused on vandalic events). For each class of events, we manually built one or more GCIs, as illustrated in Section~\ref{secrules}. The list of crime events considered is reported in Table~\ref{crimevents}. In it, the first number in parenthesis reports the number of GCIs we built for each of them, while the second number indicates the number of event instances (individuals) we  created during the manual video annotation process. So, for instance, for the event $\cp{DamageStructure}$ we have built 3 classification GCIs and we have created 9 instances of $\cp{DamageStructure}$ during the manual video annotation process. For further clarification, the 3 GCIs for  $\cp{DamageStructure}$ are
\[
\begin{array}{l}
\begin{array}{l}
\cp{Perdurant}  \ \sqcap \\
\ \exists \role{participant}.(\cp{Structure}  \ \sqcap \\
\hspace*{16ex}   \exists {\role{participantIin}}. \cp{Kicking})  \impc \cp{DamageStructure}\\
\end{array} \\
\begin{array}{l}
\cp{Perdurant}  \ \sqcap \\
\ \exists \role{participant}.(\cp{Structure}  \ \sqcap \\
\hspace*{16ex}   \exists {\role{participantIin}}. \cp{Beating})  \impc \cp{DamageStructure}\\
\end{array} \\
\begin{array}{l}
\cp{Perdurant}  \ \sqcap \\
\ \exists \role{participant}.(\cp{Structure}  \ \sqcap \\
\hspace*{16ex}   \exists {\role{participantIin}}. \cp{BreakingWindows})  \impc \cp{DamageStructure} \ ,
\end{array}
\end{array}
\]

\nd while, \eg, an instance of $\cp{DamageStructure}$ is the individual $\ind{Kicking1}$, whose related information excerpt is:
\[
\begin{array}{l}
\cass{\ind{Kicking1}}{\cp{Kicking}}, \rass{\ind{Kicking1}}{\ind{2bdf}}{\role{isFrom}}, \cass{\ind{2bdf}}{\cp{Resource}}, \rass{\ind{2bdf}}{\ind{C004}}{\role{isFrom}}, \\
\rass{\ind{2bdf}}{\ind{pr11}}{\role{has}}, \rass{\ind{pr11}}{\ind{Kicking1}}{\role{part}}, \rass{\ind{pr11}}{\ind{BreackingWindows3}}{\role{part}}, \\
\cass{\ind{BreackingWindows3}}{\cp{BreackingWindows}},   \ldots
\end{array}
\]


\nd As a matter of general information, the global metric statistics of the so built ontology is reported in Table~\ref{tab:metric}.


\vspace*{1ex}
\nd {\bf Evaluation.}
%
Let $\O$ be the built ontology from which we drop axioms stating explicitly that an individual is an instance of a crime event listed in Table~\ref{crimevents}.
Please note that without the GCIs none of the crime events instances in $\O$ can be inferred to be instances of the crime events in Table~\ref{crimevents}.\footnote{Roughly, crime events are subclasses of the  $\cp{Event}$ class, while crime event instances are instances of the class $\cp{Stative}$ (see Figure~\ref{fig:dolce2}).}

Now, on $\O$ we run an OWL 2 reasoner that determines the instances of all crime event classes in the ontology. 

To determine the classification effectiveness of the GCIs, we compute the so-called micro/macro averages of precision, recall and F1-score \wrt~inferred data, which for clarity and completeness we report below. Specifically, we measure the outcomes of binary classification using a $2 \times 2$ \emph{contingency table} or \emph{confusion matrix}, as follows:
for each  class $C$ in Table~\ref{crimevents}, consider the following contingency table

\begin{center}
\begin{tabular}{|c|c|c|c|} 
\cline{3-4}  \multicolumn{2}{c | }{} & \multicolumn{2}{c | }{True Condition} \\ 
\cline{3-4}  \multicolumn{2}{c | }{} & in $C$  &  is not in $C$ \\ \hline
\multirow{2}{*}{Prediction} &  in $C$ & $TP_C$ & $FP_C$ \\  
\cline{3-4}  & is not in $C$ & $FN_C$ & $TN_C$ \\ \hline
\end{tabular}
\end{center}

\nd where,

\begin{itemize}
\item the \emph{true positives} $TP_C$ is calculated as the number of positive predictions for the class $C$ provided  that the actual value is positive;
\item the \emph{false positives} $FP_C$ is calculated as the number of positive predictions for the class $C$  provided that the actual value is negative;
\item the \emph{false negatives} $FN_C$ is calculated as the number of negative predictions for the class $C$ provided that the actual value is positive;
\item the \emph{true negatives} $TN_C$ is calculated as the number of negative predictions for the class $C$ provided that the actual value is negative.
\end{itemize}

\nd Formally,
for each crime event class $C$, let  $true(C)$ be the set of manually determined instances $e$ of it, \ie~$e \in true(C)$  denotes that fact that $e$ has to be an instance of $C$. Then\footnote{Recall that $\O \models \cass{e}{C}$ dictates that $e$ has been automatically classified/inferred as being instance of class $C$.}
%
\begin{eqnarray*}
TP_C  & =  & \{e \mid \O \models \cass{e}{C} \mbox{ and } e \in true(C) \}  \\
FP_C  & =  &  \{e \mid \O \models \cass{e}{C} \mbox{ but } e \not \in true(C) \}  \\
FN_C  & =  &  \{e \mid \O \not\models \cass{e}{C} \mbox{ but } e \in true(C) \}  \\
TN_C  & =  &  \{e \mid \O \not\models \cass{e}{C} \mbox{ and } e \not \in true(C) \} \ .
\end{eqnarray*}
\nd Thereafter, with $|C| = |TP_C| + |FP_C|$, $|true(C)| = | TP_C | + | FN_C|$, we do the following:
\begin{enumerate}
\item for each class $C$ in Table~\ref{crimevents}, we  determine
\begin{eqnarray*}
Precision_C & = & \frac{| TP_C |}{|C|} \\
Recall_C & = & \frac{| TP_C |}{|true(C)|} \\
F1_C & = & 2\cdot \frac{Precision_C \cdot Recall_C}{Precision_C + Recall_C}
\end{eqnarray*}
\item for each of these measures we also compute the micro- and macro-average (where $N$ is the number of classes in Table~\ref{crimevents}):
\begin{eqnarray*}
Precision_{micro} & = & \frac{\sum_C | TP_C |}{\sum_C |C|} \\
Recall_{micro} & = & \frac{\sum_C | TP_C |}{\sum_C |true(C)|} \\
F1_{micro} & = & 2\cdot \frac{Precision_{micro} \cdot Recall_{micro}}{Precision_{micro} + Recall_{micro}}
\\ \\
Precision_{macro} & = & \frac{\sum_C Precision_C}{N} \\
Recall_{macro} & = &  \frac{\sum_C Recall_C}{N} \\
F1_{macro} & = & 2\cdot \frac{Precision_{macro} \cdot Recall_{macro}}{Precision_{macro} + Recall_{macro}}
\end{eqnarray*}
\end{enumerate}
The evaluation result of the first test is shown in Table~\ref{manualGCI}.

\begin{table*}[]
\centering
\caption{Results for the experiment on classification of manually build GCIs .}
\label{manualGCI}
{\large
\resizebox{\textwidth}{!}{    
\begin{tabular}{llllllllll}
    
&  &  &  &  &  &  &  &  & \\ 
\hline\hline         
\textbf    {Event} & TP & FP & FN & TN & $\left | C  \right |$ & $\left | trueC  \right |$ & $Precision_C$ &  $Recall_C$ & $F1_C$ \\ [0.5ex] 
\hline
%
%
%
%
%
%
%
%

Vandalism & 42 & 0 & 15 & 168 & 42 & 57 & 1.00 & 0.74 & \textcolor{blue}{0.85}  \\

DamageVehicle & 11 & 0 & 5 & 209 & 11 & 16 & 1.00 & 0.69 &  \textcolor{blue}{0.81} \\

DamageStructure & 9 & 0 & 0 & 216 & 9 & 9 & 0.89 & 0.89 & \textcolor{blue}{0.89}  \\

Crowding & 60 & 1 & 4 & 160 & 61 & 64 & 0.98 & 0.94 & \textcolor{blue}{ 0.96} \\

Throwing & 30 & 0 & 0 & 195 & 30 & 30 & 1.00 & 1.00 &  \textcolor{blue}{1.00} \\

Riot & 5 & 0 & 16 & 204 & 5 & 21 & 1.00 & 0.24 &  \textcolor{blue}{0.38} \\

AbnormalBehaviour & 70 & 22 & 10 & 123 & 92 & 80 & 0.76 & 0.88 & \textcolor{blue}{0.81}  \\     \hline
               
%
        & $Precision_{micro}$ & $Recall_{micro}$ & $F1_{micro}$ & $Precision_{macro}$ & $Recall_{macro}$ & $F1_{macro}$ &  &  &  \\
        &\textcolor{blue} {0.91} & \textcolor{blue} {0.82}  &\textcolor{blue}{0.86}  &\textcolor{blue} {0.96}  &\textcolor{blue} {0.78}  & \textcolor{blue} {0.86} &  &  &  \\
        \hline
        \end{tabular}%
}}    
\end{table*}

\subsection{Classification via Automatically Learned GCIs} \label{valleargci}

In the second experiment, we apply a concept learning approach to replace the manually built GCIs describing the crime events listed in Table~\ref{crimevents}.
To this end,  the DL-Learner\footnote{http://dl-learner.org/} system was used to find descriptions of the criminal events in Table~\ref{crimevents}, based on existing instances of these classes. We recall roughly that DL-learner is, among others, a framework for supervised Machine Learning in OWL~2, which is capable of making class expression suggestions for a specified class $C$ by relying on the instances of $C$.

\textbf{Setup}.  Let now  $\O$ be the ontology as in Section~\ref{valgci}, but from which we also drop additionally the manually created GCIs for the crime event listed in Table~\ref{crimevents}. 
On it we used the CELOE algorithm~\cite{buhmann2016dl,lehmann2011class} with its default settings to generate suggestion definitions (inclusion axioms) for each target class $C$ by using the manually identified crime event instances.

Specifically, we used a $K$-fold cross style validation method~\cite{forman2010apples}, which divides the available crime event instances into a $K$ disjoint subsets.
 %
 %
That is, we split each target class $C$ into $K$ disjoint subsets $C_1, \ldots, C_K$ of equal size. In our experiment, $K$ is the number of instances of $C$ and, thus, each $C_i$ has size one. 
%
%
For each $C_i$, the \emph{training set} is $(\bigcup_{i=1}^{K} C_i )\setminus C_{i}$ and is denoted as $Trainset_i$. Then, for each $C_i$ we run CELOE on the training set $Trainset_i$ and generated at most 10 class expressions of the form $D_j \impc C_i$, out of which we have chosen the best solution (denoted $D_{C_i} \impc C_i$ or $GCI_i$). If the best solution is not unique,  we select the first listed one.


The best-selected  GCIs found by CELOE for each of the target classes in Table~\ref{crimevents} are:
    \[
    \begin{array}{l}
    \cp{PhysicalAggression} \sqcap \exists \role{immediateRelation}.\cp{Structure} \sqsubseteq \cp{DamageStructure}      \\\\
    \exists\role{immediateRelation}.\cp{Vehicle} \sqsubseteq \cp{DamageVehicle}\\\\
    \exists\role{immediateRelation}.\cp{Vandalism} \sqsubseteq \cp{AbnormalBehavior} \\\\
    \exists\role{immediateRelation}.\cp{Arm} \sqsubseteq \cp{Throwing}    \\\\
    \exists\role{immediateRelation}.\cp{Group} \sqsubseteq \cp{Crowding} \ .
    \end{array}
    \]
%
With the help of a reasoner, we then infer all instances in $\O$, that are not in $Trainset_i$, that are instances of the selected $D_{C_i}$ and consider them as our \emph{result set} (denoted $Resultset_i$). 

\vspace*{1ex}
\nd \textbf{Evaluation}. 
To determine the classification effectiveness of the learned GCIs, \ie~of $GCI_i$,  precision ($Pr_i$) and recall ($Re_i$) across the folds is computed:

\begin{itemize}
    
\item the \emph{false positives} $FP_i$ is calculated as the difference between the instances in $Resultset_i$ and those in $true(C)$; 
    
\item the \emph{true positives} $TP_i$ is calculated as the difference between  $Resultset_i$ and the false positive;
    
\item the \emph{false negatives} $FN_i$ is calculated as the difference between $C_i$ and $Resultset_i$.
    
\end{itemize} 

\nd Using $FP_i, TP_i$ and $FN_i$, we have that $Pr_i = \frac{|TP_i|}{|TP_i| + |FP_i|}$, while $Re_i = \frac{|TP_i|}{|TP_i| + |FN_i|}$. We then determine  both average precision ($Precision_C$) and recall ($Recall_C$) across the folds, \ie
%
\begin{eqnarray*}
Precision_C = \frac{1}{K}\cdot\sum_{i=1}^{k} Pr_i\\
Recall_C = \frac{1}{K}\cdot \sum_{i=1}^{k} Re_i \ .
\end{eqnarray*}
    
\begin{table*}
        \caption{Results for the experiment on classification using DL-Learner CELOE algorithm.}        \label{DL-Learner}        
\begin{center}
            \begin{tabular}{lllll} 
                &  &  &    \\ 
                \hline\hline 
                
                Event & $Precision_C$ &  $Recall_C$ & $F1_C$\\ [0.5ex]
                \hline
                DamageVehicle    & 0.69 & 0.98 &  \textcolor{blue}{0.81} \\ 
                
                Damage Structure & 1.00     & 1.00     &   \textcolor{blue}{1.00}    \\  
                
                Crowding         & 0.96 & 1.00 &  \textcolor{blue}{0.98}  \\ 
                
                Throwing         & 0.86 & 0.99 &  \textcolor{blue}{0.92}  \\  
                
                AbnormalBehavior & 0.69& 0.99 & \textcolor{blue}{0.81}   \\  \hline \\

\end{tabular}

\begin{tabular}{llllll} \hline
                $Precision_{micro}$ & $Recall_{micro}$ & $F1_{micro}$ & $Precision_{macro}$ & $Recall_{macro}$ & $F1_{macro}$    \\
\textcolor{blue}{0.753}  &  \textcolor{blue}{0.964} &\textcolor{blue}{0.845}  &\textcolor{blue}{0.599}  &\textcolor{blue} { 0.709} & \textcolor{blue}{0.649}  \\                \hline
            \end{tabular}%
\end{center}            
    \end{table*}

\nd Using these measures we then compute the micro- and macro-averages of precision, recall and F1 as in Section~\ref{valgci}.

%
    
The evaluation results of the second test are shown in Table~\ref{DL-Learner}. 

\vspace*{1ex}
\nd {\bf Discussion.}
The results are generally promising. In the manually built GCI case, precision and F1 are reasonably good, though in one case ($\cp{Riot}$) the recall and, thus, F1 is not satisfactory.
For the learned GCI case, the individual measures are generally comparable to the manual ones. 

Given that the learned GCIs are completely different than the manually built ones, it is surprising that both sets perform more or less the same. However, please note that DL-Learner was neither able to learn a GCI  for $\cp{Vandalism}$ nor for $\cp{Riot}$. This fact is reflected in the generally worse micro/macro precision, recall and F1 measures.

Eventually, we also merged the manually built GCIs and the learned ones together and tested them as in Section~\ref{valgci}. The results in Table~\ref{merge} show, however, that globally their effectiveness is as for the manual case (and does not improve).

\begin{table*}
        \caption{Results of merging manual and learned GCIs.}        \label{merge}        
\begin{center}
            \begin{tabular}{lllll} 
                &  &  &    \\ 
                \hline\hline 
                
                Event & $Precision_C$ &  $Recall_C$ & $F1_C$\\ [0.5ex]                \hline
                Vandalism    & 1.00 & 0.74 &  \textcolor{blue}{0.85} \\ 
                DamageVehicle    & 1.00 & 0.69 &  \textcolor{blue}{0.81} \\ 
                
                Damage Structure & 0.89     & 0.89     &   \textcolor{blue}{0.89}    \\  
                
                Crowding         & 0.98 & 0.94 &  \textcolor{blue}{0.96}  \\ 
                
                Throwing         & 1.00 & 1.00 &  \textcolor{blue}{1.00}  \\  
                
                Riot         & 1.00 & 0.24 &  \textcolor{blue}{0.38}  \\  
                
                AbnormalBehavior & 0.76 & 0.89 & \textcolor{blue}{0.82}   \\  \hline \\

\end{tabular}

\begin{tabular}{llllll} \hline
                $Precision_{micro}$ & $Recall_{micro}$ & $F1_{micro}$ & $Precision_{macro}$ & $Recall_{macro}$ & $F1_{macro}$    \\
\textcolor{blue}{0.90}  &  \textcolor{blue}{0.82} &\textcolor{blue}{0.86}  &\textcolor{blue}{0.95}  &\textcolor{blue} { 0.77} & \textcolor{blue}{0.85}  \\                \hline
            \end{tabular}%
\end{center}            
    \end{table*}

\section{Conclusions}
\label{sec:Conclusion}

In this work, we have proposed an extensive ontology for representing complex criminal events. The proposed ontology focuses on events that are often required by forensic analysts. In this context, the $\cp{Perdurant}$, as defined in the DOLCE ontology as an occurrence in time, and the $\cp{Endurant}$, defined in the DOLCE ontology as contentious in time, have both been extended to represent all forensics entities together with meaningful entities for video surveillance-based vandalism detection.
The aim of the built ontology is to support the interoperability of the automated surveillance system. 



To classify high-level events in terms of the composition of lower level events we focused on both manually built and automatically learned GCIs and have compared the evaluation results of both experiments. The results are generally promising and the effectiveness of machine derived definitions for high-level crime events is encouraging though needs further development.

In the future, we intend to deal with vague or imprecise knowledge and we would like to work on the problem of automatically learn fuzzy concept description~\cite{Bobillo11c,Bobillo16,Lisi13,Lisi14,Lukasiewicz08a,Straccia13,Straccia15} as mosto of the involved entities are fuzzy.

\section*{Acknowledgements}
This work is partially funded by the European Union's Seventh Framework Programme, grant agreement number 607480 (LASIE IP project).


\begin{thebibliography}{10}

\bibitem{appan2004networked}
Preetha Appan and Hari Sundaram.
\newblock Networked multimedia event exploration.
\newblock In {\em Proceedings of the 12th Annual ACM International Conference
  on Multimedia}, pages 40--47. ACM, 2004.

\bibitem{Baader03a}
Franz Baader, Diego Calvanese, Deborah McGuinness, Daniele Nardi, and Peter~F.
  Patel-Schneider, editors.
\newblock {\em The Description Logic Handbook: Theory, Implementation, and
  Applications}.
\newblock Cambridge University Press, 2003.

\bibitem{Baader09}
Franz Baader, Ian Horrocks, and Ulrike Sattler.
\newblock Description logics.
\newblock In Steffen Staab and Rudi Studer, editors, {\em Handbook on
  Ontologies}, International Handbooks on Information Systems, pages 21--43.
  Springer Verlag, 2009.

\bibitem{Bobillo11c}
Fernando Bobillo and Umberto Straccia.
\newblock Fuzzy ontology representation using {OWL} 2.
\newblock {\em International Journal of Approximate Reasoning}, 52:1073--1094,
  2011.

\bibitem{Bobillo16}
Fernando Bobillo and Umberto Straccia.
\newblock The fuzzy ontology reasoner \emph{{fuzzyDL}}.
\newblock {\em Knowledge-Based Systems}, 95:12 -- 34, 2016.

\bibitem{buhmann2016dl}
Lorenz B{\"u}hmann, Jens Lehmann, and Patrick Westphal.
\newblock {DL}-{L}earner framework for inductive learning on the semantic web.
\newblock {\em Web Semantics: Science, Services and Agents on the World Wide
  Web}, 39:15--24, 2016.

\bibitem{sep-events}
Roberto Casati and Achille Varzi.
\newblock Events.
\newblock In Edward~N. Zalta, editor, {\em The Stanford Encyclopedia of
  Philosophy}. Winter 2015 edition, 2015.
\newblock \url{http://plato.stanford.edu/archives/win2015/entries/events/}.

\bibitem{forman2010apples}
George Forman and Martin Scholz.
\newblock Apples-to-apples in cross-validation studies: pitfalls in classifier
  performance measurement.
\newblock {\em ACM SIGKDD Explorations Newsletter}, 12(1):49--57, 2010.

\bibitem{francois2005verl}
Alexandre R.~J. Fran{\c{c}}ois, Ram Nevatia, Jerry~R. Hobbs, and Robert~C.
  Bolles.
\newblock {VERL:} an ontology framework for representing and annotating video
  events.
\newblock {\em {IEEE} MultiMedia}, 12(4):76--86, 2005.

\bibitem{hakeem2005object}
Asaad Hakeem, Khurram Shafique, and Mubarak Shah.
\newblock An object-based video coding framework for video sequences obtained
  from static cameras.
\newblock In {\em Proceedings of the 13th Annual ACM International Conference
  on Multimedia}, pages 608--617. ACM, 2005.

\bibitem{Hakeem04}
Asaad Hakeem, Yaser Sheikh, and Mubarak Shah.
\newblock {CASE$^E$}: A hierarchical event representation for the analysis of
  videos.
\newblock In {\em Proceedings of the 19th National Conference on Artificial
  Intelligence}, pages 263--268. AAAI Press, 2004.

\bibitem{Henderson15}
Craig Henderson, Saverio~G. Blasi, Faranak Sobhani, , and Ebroul Izquierdo.
\newblock On the impurity of street-scene video footage.
\newblock In {\em 6th International Conference on Imaging for Crime Prevention
  and Detection (ICDP-15)}. The Institution of Engineering and Technology
  (IET), 2015.

\bibitem{jain2003experiential}
Ramesh Jain, Pilho Kim, and Zhao Li.
\newblock Experiential meeting system.
\newblock In {\em Proceedings of the 2003 ACM SIGMM Workshop on Experiential
  Telepresence}, pages 1--12. ACM, 2003.

\bibitem{kim2004personal}
Pilho Kim, Mark Podlaseck, and Gopal Pingali.
\newblock Personal chronicling tools for enhancing information archival and
  collaboration in enterprises.
\newblock In {\em Proceedings of the the 1st ACM Workshop on Continuous
  Archival and Retrieval of Personal Experiences}, pages 56--65. ACM, 2004.

\bibitem{lehmann2011class}
Jens Lehmann, S{\"o}ren Auer, Lorenz B{\"u}hmann, and Sebastian Tramp.
\newblock Class expression learning for ontology engineering.
\newblock {\em Web Semantics: Science, Services and Agents on the World Wide
  Web}, 9(1):71--81, 2011.

\bibitem{Lisi13}
Francesca~A. Lisi and Umberto Straccia.
\newblock A logic-based computational method for the automated induction of
  fuzzy ontology axioms.
\newblock {\em Fundamenta Informaticae}, 124(4):503--519, 2013.

\bibitem{Lisi14}
Francesca~Alessandra Lisi and Umberto Straccia.
\newblock A foil-like method for learning under incompleteness and vagueness.
\newblock In {\em 23rd International Conference on Inductive Logic
  Programming}, volume 8812 of {\em Lecture Notes in Artificial Intelligence},
  pages 123--139, Berlin, 2014. Springer Verlag.
\newblock Revised Selected Papers.

\bibitem{Lukasiewicz08a}
Thomas Lukasiewicz and Umberto Straccia.
\newblock Managing uncertainty and vagueness in description logics for the
  semantic web.
\newblock {\em Journal of Web Semantics}, 6:291--308, 2008.

\bibitem{masolo2003wonderweb}
Claudio Masolo, Stefano Borgo, Aldo Gangemi, Nicola Guarino, and Alessandro
  Oltramari.
\newblock Wonderweb {D}eliverable {D18}, {O}ntology {L}ibrary (final).
\newblock {\em ICT project}, 33052, 2003.

\bibitem{Meghini01}
Carlo Meghini, Fabrizio Sebastiani, and Umberto Straccia.
\newblock A model of multimedia information retrieval.
\newblock {\em Journal of the ACM}, 48(5):909--970, 2001.

\bibitem{motik2009owl}
Boris Motik, Bernardo~Cuenca Grau, Ian Horrocks, Zhe Wu, Achille Fokoue,
  Carsten Lutz, et~al.
\newblock {OWL 2} {W}eb {O}ntology {L}anguage {P}rofiles.
\newblock {\em W3C recommendation}, 27:61, 2009.

\bibitem{nevatia2004ontology}
Ram Nevatia, Jerry~R. Hobbs, and Bob Bolles.
\newblock An ontology for video event representation.
\newblock In {\em {IEEE} Conference on Computer Vision and Pattern Recognition
  Workshops}, page 119. {IEEE} Computer Society, 2004.

\bibitem{nevatia2003hierarchical}
Ram Nevatia, Tao Zhao, and Somboon Hongeng.
\newblock Hierarchical language-based representation of events in video
  streams.
\newblock In {\em {IEEE} Conference on Computer Vision and Pattern
  Recognition}, page~39. {IEEE} Computer Society, 2003.

\bibitem{SWRL}
{SWRL}: A Semantic Web Rule Language~Combining OWL and RuleML.
\newblock {\em \url{https://www.w3.org/Submission/SWRL/}}.
\newblock {W3C}, 2004.

\bibitem{OWL2}
{\mbox{{OWL 2 Web Ontology Language} Document Overview}}.
\newblock {\em \url{http://www.w3.org/TR/2009/REC-owl2-overview-20091027/}}.
\newblock {W3C}, 2009.

\bibitem{pingali2002instantly}
Gopal~Sarma Pingali, Agata Opalach, Yves~D. Jean, and Ingrid~B. Carlbom.
\newblock Instantly indexed multimedia databases of real world events.
\newblock {\em IEEE Transactions on Multimedia}, 4(2):269--282, 2002.

\bibitem{rothstein2004verb}
Susan Rothstein.
\newblock Chapter 1: Verb classes and aspectual classification.
\newblock In {\em Structuring Events: A Study in the Semantics of Lexical
  Aspect}, pages 1--35. Wiley Online Library, 2004.

\bibitem{scherp2009f}
Ansgar Scherp, Thomas Franz, Carsten Saathoff, and Steffen Staab.
\newblock F--{A} model of events based on the foundational ontology {DOLCE+DnS}
  ultralight.
\newblock In {\em Proceedings of the Fifth International Conference on
  Knowledge Capture}, pages 137--144. ACM, 2009.

\bibitem{Schmidt-Schauss91}
Manfred Schmidt-Schau{\ss} and Gert Smolka.
\newblock Attributive concept descriptions with complements.
\newblock {\em Artificial Intelligence}, 48:1--26, 1991.

\bibitem{snidaro2007representing}
Lauro Snidaro, Massimo Belluz, and Gian~Luca Foresti.
\newblock Representing and recognizing complex events in surveillance
  applications.
\newblock In {\em Fourth {IEEE} International Conference on Advanced Video and
  Signal Based Surveillance}, pages 493--498. {IEEE} Computer Society, 2007.

\bibitem{Sobhani16}
Faranak Sobhani, Krishna Chandramouli, Qianni Zhang, and Ebroul Izquierdo.
\newblock Formal representation of events in a surveillance domain ontology.
\newblock In {\em 2016 {IEEE} International Conference on Image Processing},
  pages 913--917. {IEEE} Computer Society, 2016.

\bibitem{Sobhani15}
Faranak Sobhani, Nur~Farhan Kahar, and Qianni Zhang.
\newblock An ontology framework for automated visual surveillance system.
\newblock In {\em 13th International Workshop on Content-Based Multimedia
  Indexing}, pages 1--7. {IEEE} Computer Society, 2015.

\bibitem{Straccia13}
Umberto Straccia.
\newblock {\em Foundations of Fuzzy Logic and Semantic Web Languages}.
\newblock CRC Studies in Informatics Series. Chapman {\&} Hall, 2013.

\bibitem{Straccia15}
Umberto Straccia and Matteo Mucci.
\newblock {pFOIL-DL}: Learning (fuzzy) $\mathcal{EL}$ concept descriptions from
  crisp {OWL} data using a probabilistic ensemble estimation.
\newblock In {\em Proceedings of the 30th Annual ACM Symposium on Applied
  Computing (SAC-15)}, pages 345--352, Salamanca, Spain, 2015. ACM.

\bibitem{vendler1957verbs}
Zeno Vendler.
\newblock Verbs and times.
\newblock {\em The Philosophical Review}, 62(2):143--160, 1957.

\bibitem{vendler1967linguistics}
Zeno Vendler, editor.
\newblock {\em Linguistics in Philosophy}.
\newblock G - Reference, Information and Interdisciplinary Subjects Series.
  Cornell University Press, 1967.

\bibitem{westermann2007toward}
Utz Westermann and Ramesh Jain.
\newblock Toward a common event model for multimedia applications.
\newblock {\em IEEE Multimedia}, 14(1), 2007.

\end{thebibliography}
\end{document}